\pdfoutput=1
\documentclass[10pt,twocolumn,letterpaper]{article}

\usepackage[pagenumbers]{cvpr}

\usepackage{booktabs}
\usepackage{colortbl}
\usepackage{tabularx}
\usepackage{pgfplots}
\usepackage{pgfplotstable}
\pgfplotsset{compat=1.18}
\usetikzlibrary{arrows.meta,calc}
\usepackage{enumitem}
\usepackage[super]{nth}

\newcommand{\intrinsiclabelrow}{%
  \makebox[\linewidth]{\small%
    \makebox[0.1666\linewidth]{Input}%
    \makebox[0.1666\linewidth]{Albedo}%
    \makebox[0.1666\linewidth]{Irradiance}%
    \makebox[0.1666\linewidth]{Normals}%
    \makebox[0.1666\linewidth]{Roughness}%
    \makebox[0.1666\linewidth]{Metallicity}}}
\newcommand{\myparagraph}[1]{\vspace{4pt}\noindent\textbf{#1}}

\definecolor{cvprblue}{rgb}{0.21,0.49,0.74}
\usepackage[pagebackref,breaklinks,colorlinks,allcolors=cvprblue]{hyperref}

\newcommand{\methodName}{RAID}

\title{\textbf{\methodName:} Representation Autoencoder for Intrinsic Decomposition}

\author{
Claudia Cuttano$^{1,2}$\footnotemark[1]
\quad
Tamy Boubekeur$^{1}$
\quad
Elena Garces$^{1}$
\quad
Giuseppe Vecchio$^{1}$\\
{\tt\small \{ccuttano, boubek, elenag, gvecchio\}@adobe.com}\\
$^{1}$Adobe Research, France
\quad
$^{2}$Politecnico di Torino, Italy
}

\begin{document}
\twocolumn[{%
\renewcommand\twocolumn[1][]{#1}%
\maketitle
\centering

\begin{minipage}{\textwidth}
    \centering
    \intrinsiclabelrow\par\vspace{1pt}
    \includegraphics[width=\linewidth]{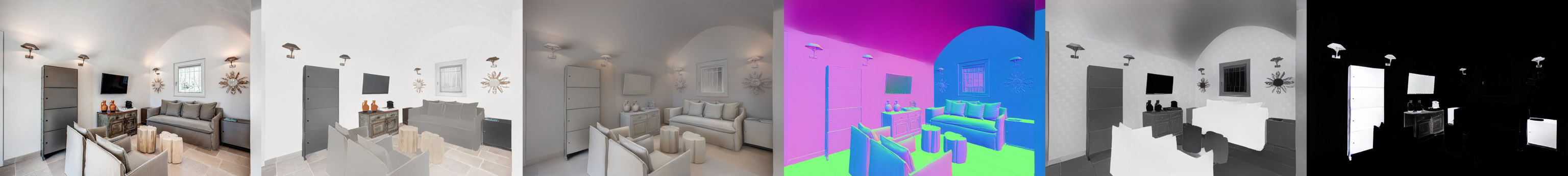}
    \vspace{0.4mm}
    \includegraphics[width=\linewidth]{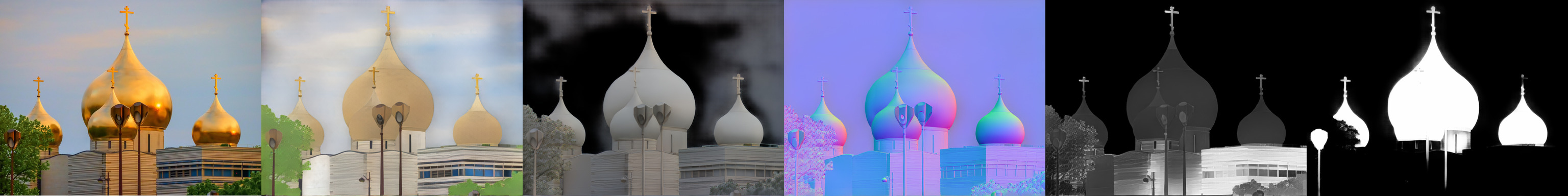}
     \vspace{0.4mm}
    \includegraphics[width=\linewidth]{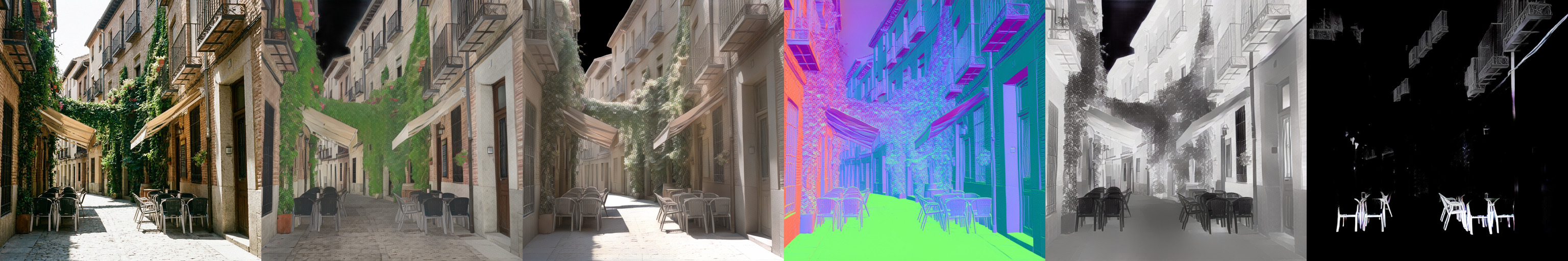}
    \vspace{0.4mm}
    \includegraphics[width=\linewidth]{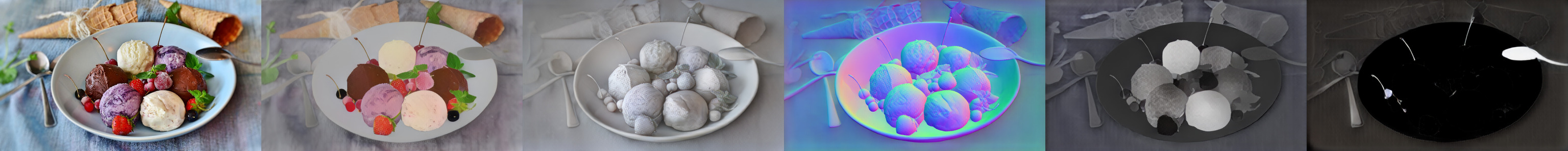}
\end{minipage}

\captionof{figure}{
We propose \textbf{\methodName{}}, a \textbf{R}epresentation \textbf{A}utoencoder for single-image \textbf{I}ntrinsic \textbf{D}ecomposition that learns to reconstruct intrinsic maps before learning to infer them from RGB.
A single forward pass predicts reflectance (albedo), illumination (irradiance), geometry (surface normals), and material properties (roughness and metallicity). \methodName{} achieves over $100{\times}$ the throughput of diffusion baselines with a fraction of their parameters, while maintaining stable predictions as input resolution varies.
\label{fig:teaser}}
\vspace{1.3\baselineskip}
}]

\footnotetext[1]{Work done during the internship.}

\begin{abstract}
Single-image intrinsic decomposition is an inherently ill-posed problem, requiring the disentanglement of strongly coupled appearance factors from a single RGB observation. Recent methods address this ambiguity with diffusion priors, at the cost of large, iterative, and often stochastic models. We show that accurate decomposition does not require a large generative model. Our key idea is to decouple reconstruction from decomposition on top of a strong pretrained representation. First, an $X \rightarrow X$ teacher learns to reconstruct each intrinsic map from DINOv3 features using a shared decoder, without paired RGB-to-intrinsic supervision.
Then, a DINOv3-based student predicts the corresponding teacher representations from RGB through lightweight, modality-specific projections, enabling training on partially annotated datasets and predicting all intrinsics in a single forward pass.
\textbf{\methodName{}}, predicts intrinsic properties in the structured latent space, enabling a compact 165M-parameter model, 100× faster than diffusion-based methods.
Despite its compact size, it achieves competitive or superior decomposition quality, with improved robustness to input resolution. Additionally, our learned projections yield interpretable, modality-specific feature spaces. 
\end{abstract}

\section{Introduction}
\label{sec:intro}

Recovering the physical properties of a scene from visual observations is a long-standing problem in computer vision and graphics~\cite{garces2022survey}. Given an RGB image, intrinsic decomposition aims to 
recover the underlying properties that determine the observed appearance, such as albedo, surface normals, illumination, roughness, and metallicity, as shown in \cref{fig:teaser}. 
These quantities underpin physically-based rendering pipelines: albedo and material parameters determine surface reflectance, normals describe geometry, and irradiance captures illumination. Reliable and efficient estimation of such maps would enable applications in image editing, relighting, and compositing~\cite{Zeng:2024:rgbx, Careaga:2025:Relighting, ren:2025:mvcolight}, asset creation and PBR material generation~\cite{Xiong:2025:TexGaussian}, and physical scene understanding, simulation, and interaction~\cite{Roberts:2021:Hypersim, Zhai:2024:PhysicalProperty, Cai:2024:GIC, Xu:2025:GaussianProperty}.

The problem is inherently ambiguous: a single RGB observation can be explained by many combinations of material, lighting, and geometry. A dark region, for instance, may be a low-reflectance surface under strong illumination, a high-reflectance surface under weak illumination, or a surface oriented away from the light. A model must therefore address several challenges at once: disentangling these factors, which is especially hard for roughness and metallicity; reconstructing them as sharp, accurate maps; and learning both from limited supervision.
This calls for both high-level semantic understanding of materials and fine-grained spatial information.

To this end, recent methods~\cite{Zeng:2024:rgbx, Dirik:2026:PRISM, dirik:2026:geoid} exploit the generative prior of image diffusion models, which constrains the ill-posed inverse problem when the denoising process is conditioned on the RGB input. However, this typically requires multiple stochastic denoising steps per intrinsic map, incurring substantial computational and memory costs, particularly for multiple properties at high resolution. Even single-step variants~\cite{He:2025:Lotus, Pavlovic:2026:MarigoldV2} inherit the cost of large generative backbones. This hinders fast, repeatable, and scalable decomposition, as needed for interactive editing and large-scale image processing.

An alternative direction is to leverage pretrained visual representations as a foundation for intrinsic decomposition. Self-supervised vision models~\cite{Caron:2021:DINO,Oquab:2023:Dinov2,Simeoni:2025:Dinov3} learn transferable features that capture semantic and spatial information without task-specific supervision. DINOv3~\cite{Simeoni:2025:Dinov3} scales up this paradigm and introduces Gram anchoring to preserve the spatial structure of dense features, making it well suited to dense prediction~\cite{Cuttano:2026:INSID3}. Its semantics can inform otherwise ambiguous decisions, such as whether a surface is metallic or glossy, while its dense features retain spatial detail.

Using these features directly, however, entangles two problems in a single optimization: inferring physical properties from RGB and reconstructing them as detailed pixel-space maps. Since DINO features have proven effective as reconstruction latents and representation-alignment targets~\cite{Zheng:2026:RAE,Yu:2025:REPA}, we use the DINOv3 representation as a common latent space and decouple the two problems.

Building on this, we introduce \textbf{\methodName{}}: \textbf{R}epresentation \textbf{A}utoencoder for \textbf{I}ntrinsic \textbf{D}ecomposition. We first train an $X \rightarrow X$ autoencoder that learns a single, modality-agnostic decoder that maps the DINOv3 features of RGB, albedo, normals, irradiance, roughness, and metallicity back to pixel space, without addressing the inverse problem. We then freeze this teacher and train an $\mathrm{RGB} \rightarrow X$ student whose modality-specific projection heads predict, from the RGB features, the teacher representation of each map, which the shared decoder converts into the final map. This design also relaxes data requirements: the teacher reconstructs each map from itself and needs no paired data, and the student is supervised only on the maps each dataset provides, so partially annotated datasets can be combined.

In summary, our contributions are:
\begin{itemize}[leftmargin=*,itemsep=1pt,topsep=2pt]
\item We introduce a representation autoencoder for intrinsic decomposition, showing that physical scene properties can be inferred from DINOv3 features and reconstructed from a shared latent space. By learning pixel-space reconstruction in the autoencoder, we can train the RGB student to align with the semantic and spatial structure of each intrinsic property in feature space, decoupling the inference of physical factors from the reconstruction of high-quality pixel-level maps.

\item We achieve a strong quality--efficiency trade-off with a $165$M-parameter model: competitive or superior accuracy, particularly on roughness and metallicity, at over $100{\times}$ the throughput of diffusion baselines. Compared with the previous feed-forward baseline~\cite{wu:2025:MVInverse}, \methodName{} uses $6{\times}$ fewer parameters and runs about $1.7{\times}$ faster.

\item We demonstrate superior scalability to input resolution compared to prior methods, maintaining stable prediction quality and low memory footprint as resolution increases. We further show that an interpretable organization by physical property emerges in the learned modality projections: feature similarities reflect surface reflectance, geometry, illumination, and material properties, marking a step toward latent representations that support explicit reasoning about physical factors.
\end{itemize}

\section{Related Work}

\begin{figure*}[t]
\centering
\begingroup

\definecolor{mNavy}{HTML}{254A68}
\definecolor{mBlue}{HTML}{E9F1F8}
\definecolor{mTeal}{HTML}{2D716D}
\definecolor{mRed}{HTML}{B65353}
\definecolor{mPink}{HTML}{FBE5E3}
\definecolor{mInk}{HTML}{34434E}
\definecolor{mRule}{HTML}{BCC9D2}

\tikzset{
  model/.style={
    draw=mNavy,
    fill=mBlue,
    line width=.75pt,
    rounded corners=3pt
  },
  decoder/.style={
    draw=mRed,
    fill=mPink,
    line width=.75pt,
    rounded corners=3pt
  },
  projection/.style={
    draw=mRed!65,
    fill=mPink,
    line width=.55pt,
    rounded corners=2pt
  },
  signal/.style={
    draw=mInk,
    line width=.72pt,
    -{Stealth[length=2mm,width=1.3mm]}
  },
  alignment/.style={
    draw=mTeal,
    line width=.72pt,
    {Stealth[length=1.8mm]}-{Stealth[length=1.8mm]}
  },
  modeltext/.style={
    text=mNavy,
    font=\normalfont\bfseries\fontsize{8.3}{9.8}\selectfont
  },
  inputlabel/.style={
    text=mInk,
    font=\normalfont\fontsize{8.5}{9.5}\selectfont,
    inner xsep=5pt
  },
  alignmenttext/.style={
    align=center,
    text=mTeal,
    font=\normalfont\itshape\fontsize{8.5}{9.5}\selectfont,
    inner sep=0pt,
    outer sep=0pt
  },
  projectiontitle/.style={
    text=mInk,
    font=\normalfont\bfseries\fontsize{6.8}{8}\selectfont
  },
  modality/.style={
    text=mInk,
    font=\normalfont\fontsize{6.8}{8}\selectfont
  }
}

\def\methodFlame{%
  \includegraphics[width=0.38cm]
    {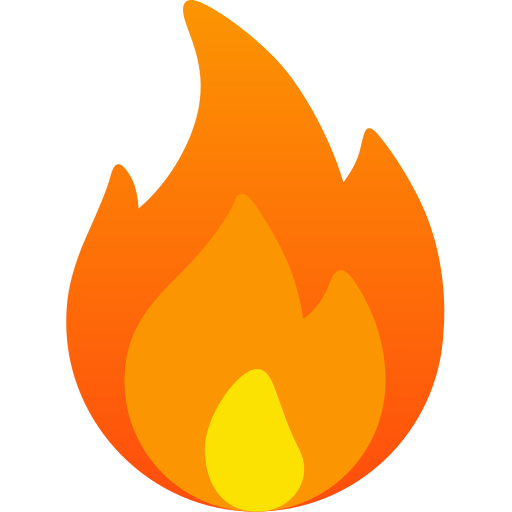}%
}

\def\methodSnowflake#1#2{%
  \begin{scope}[shift={(#1,#2)},draw=mNavy,line width=.75pt,line cap=round]
    \foreach \a in {0,60,120} {\draw (\a:-11) -- (\a:11);}
    \foreach \a in {0,60,...,300} {\draw (\a:6) -- ++(\a+45:4.5); \draw (\a:6) -- ++(\a-45:4.5);}
  \end{scope}%
}

\def\methodAdapter#1#2#3#4{%
  \begin{scope}[shift={(#1,#2)}]
    \path[draw=#3,fill=#4,line width=.65pt]
      (0,0)--(25,0)--(18,17)--(7,17)--cycle;
    \path[draw=#3,fill=#4,line width=.65pt]
      (7,17)--(18,17)--(25,34)--(0,34)--cycle;
  \end{scope}%
}

\resizebox{\textwidth}{!}{%
\begin{tikzpicture}[
  x=0.015cm,
  y=-0.015cm,
  every node/.style={inner sep=0pt,outer sep=0pt},
  line join=round
]

\path[use as bounding box]
  (175,-10) rectangle (1320,605);

\node[anchor=north west] at (266.300,191.000) {%
  \includegraphics[width=1.072cm,height=2.000cm]
    {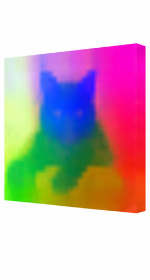}
};

\draw[signal] (302,333) -- (302,307);

\path[model] (237,333) rectangle (367,396);
\methodAdapter{220}{350}{mRed}{mPink}

\node[modeltext] at (302,364.5) {DINOv3};

\draw[signal] (302,116) -- (302,100);
\path[decoder] (237,116) rectangle (367,186);

\node[modeltext] at (302,151) {DC-AE};

\draw[signal] (302,191) -- (302,186);

\node[anchor=north west] at (255.330,26.000) {%
  \includegraphics[width=1.400cm,height=1.050cm]
    {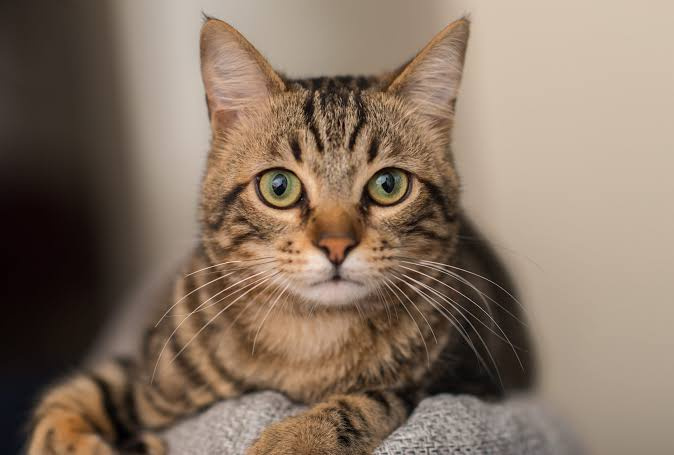}
};

\draw[
  draw=mRule,
  densely dashed,
  line width=.65pt
] (423,605) -- (423,16);

\path[model]
  (865.500,337) rectangle (982.500,398);

\methodAdapter{848.4}{350}{mRed}{mPink}

\node[modeltext] at (924.000,367.5) {DINOv3};

\draw[signal] (922.200,417) -- (922.200,399);

\node[anchor=north west] at (850.353,417.380) {%
  \includegraphics[width=2.395cm,height=1.796cm]
    {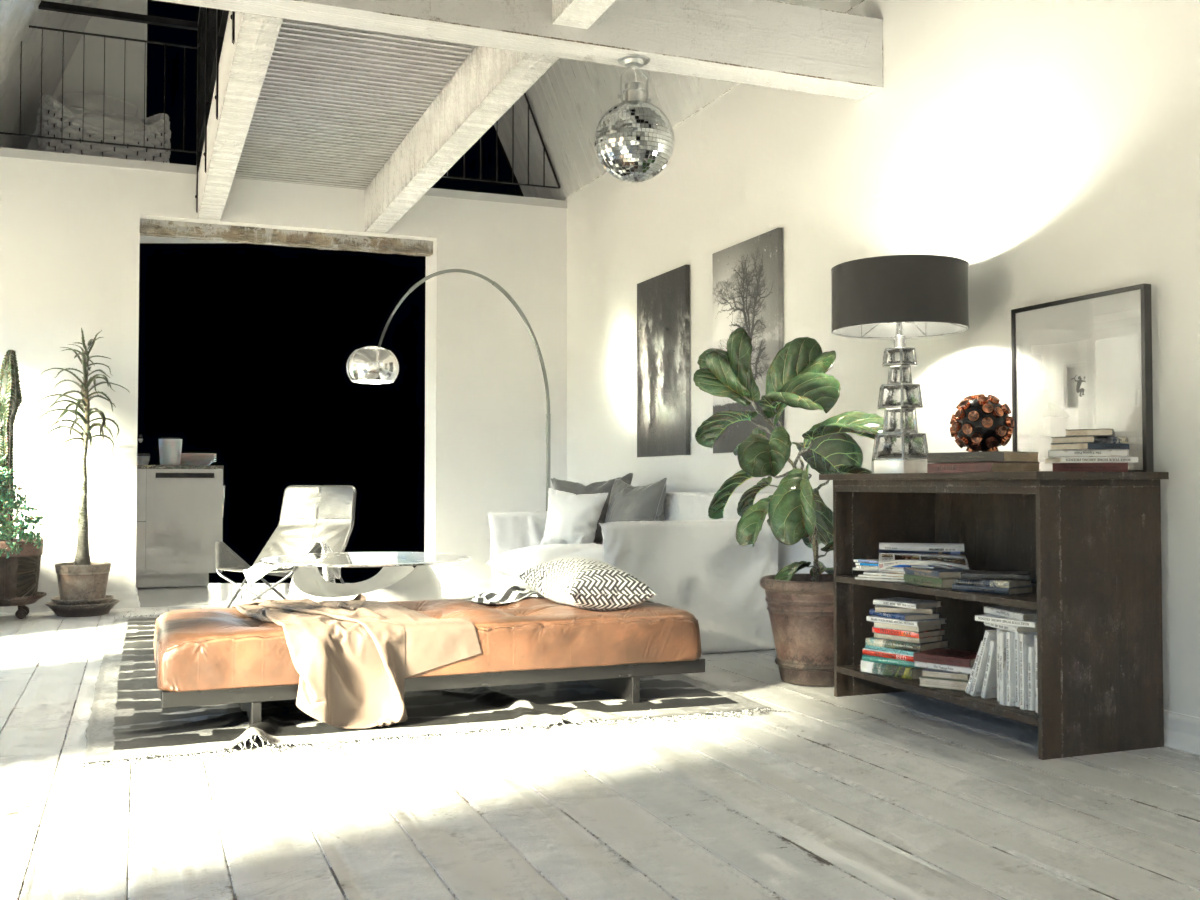}
};

\draw[signal] (531,335) -- (531,309);

\path[model] (466,335) rectangle (596,396);
\methodAdapter{449}{350}{mNavy}{mBlue}

\node[modeltext] at (531,365.5) {DINOv3};

\path[model]
  (863.700,116) rectangle (980.700,186);

\node[modeltext] at (922.200,151) {DC-AE};

\node[anchor=north west] at (488.275,419.186) {%
  \includegraphics[width=2.049cm,height=1.537cm]
    {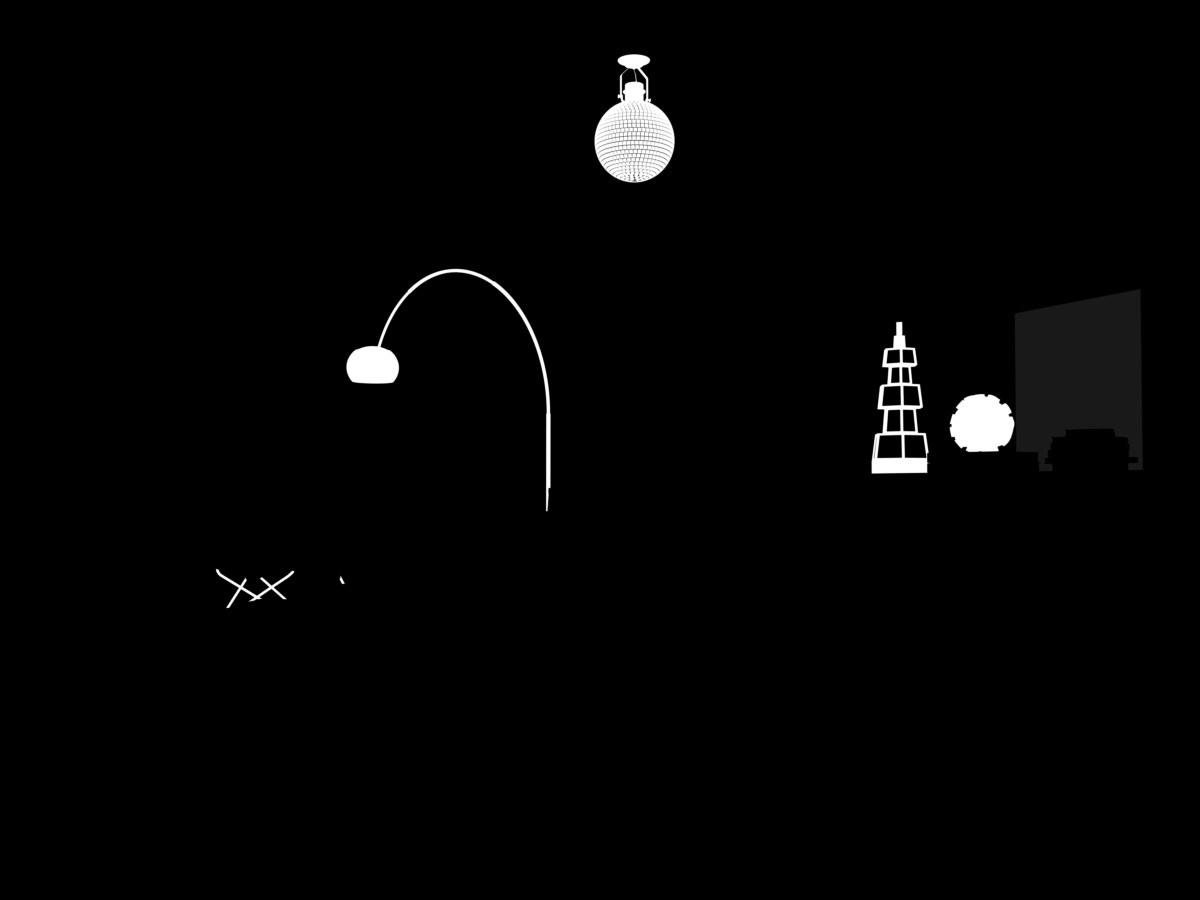}
};
\node[anchor=north west] at (481.332,425.558) {%
  \includegraphics[width=2.049cm,height=1.537cm]
    {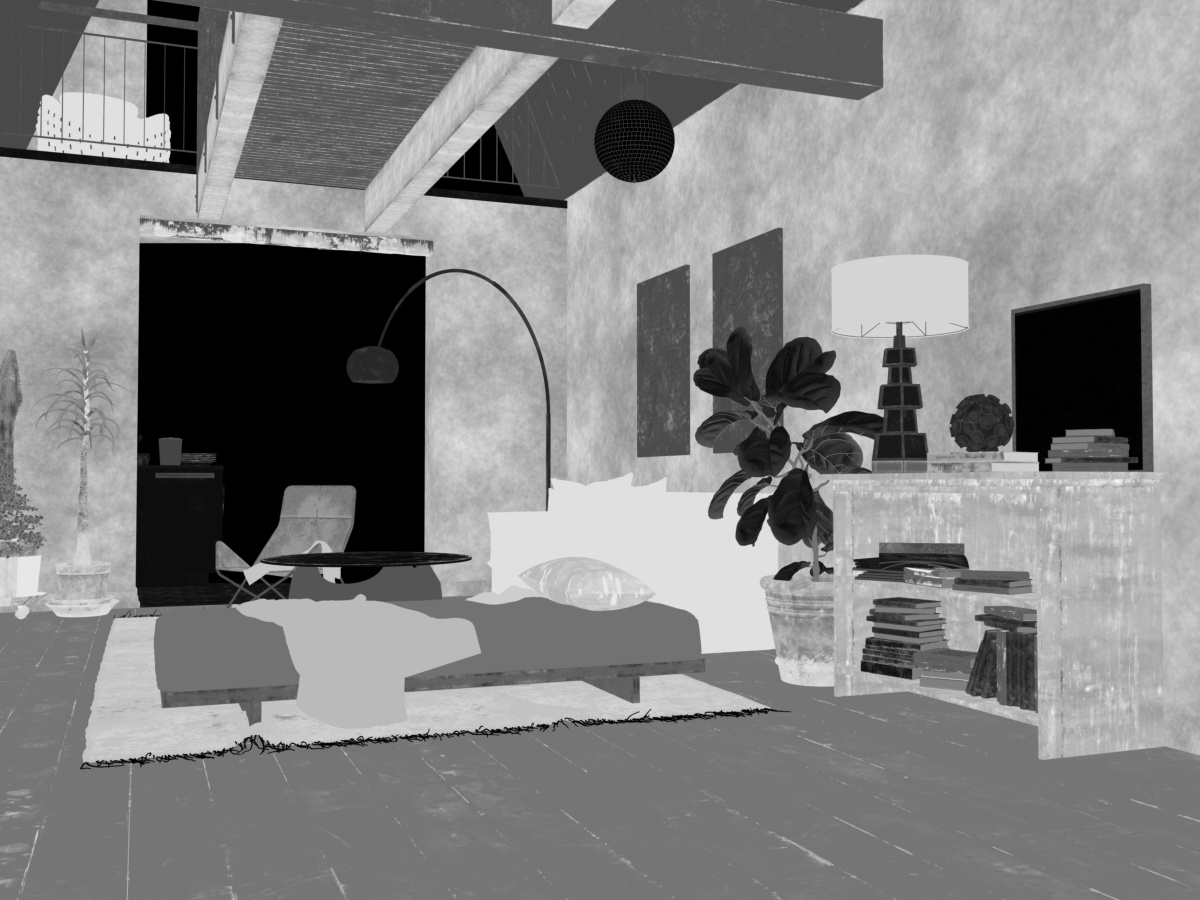}
};
\node[anchor=north west] at (474.894,430.337) {%
  \includegraphics[width=2.049cm,height=1.537cm]
    {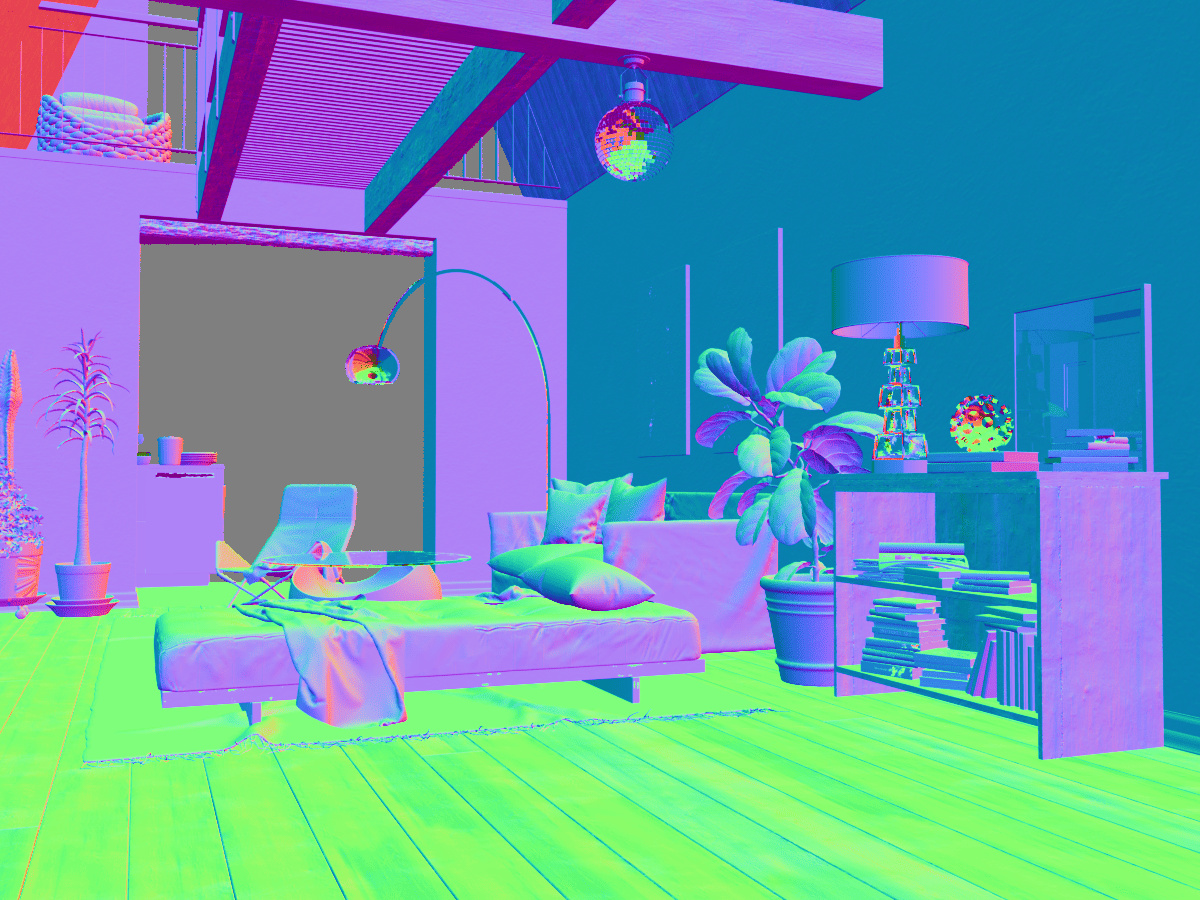}
};
\node[anchor=north west] at (468.947,436.709) {%
  \includegraphics[width=2.049cm,height=1.537cm]
    {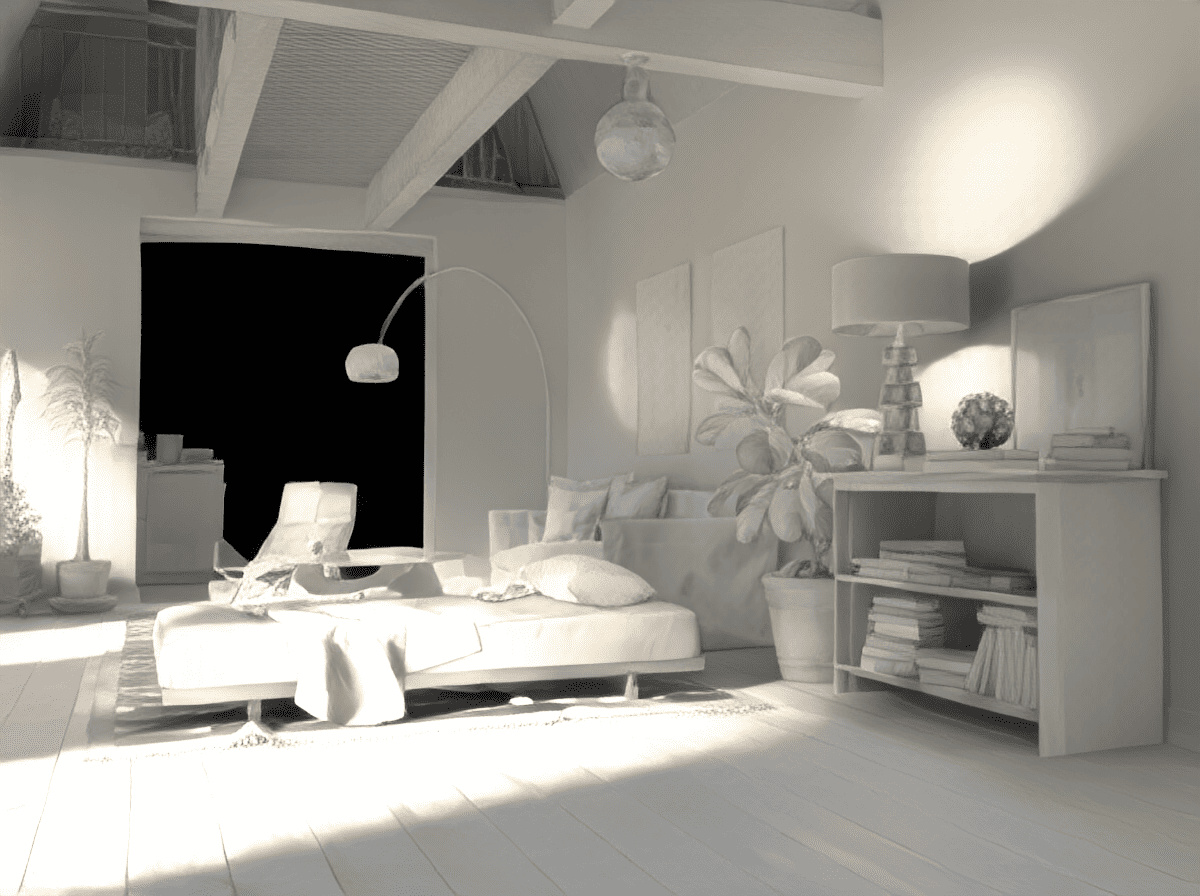}
};
\node[anchor=north west] at (463.000,441.488) {%
  \includegraphics[width=2.049cm,height=1.537cm]
    {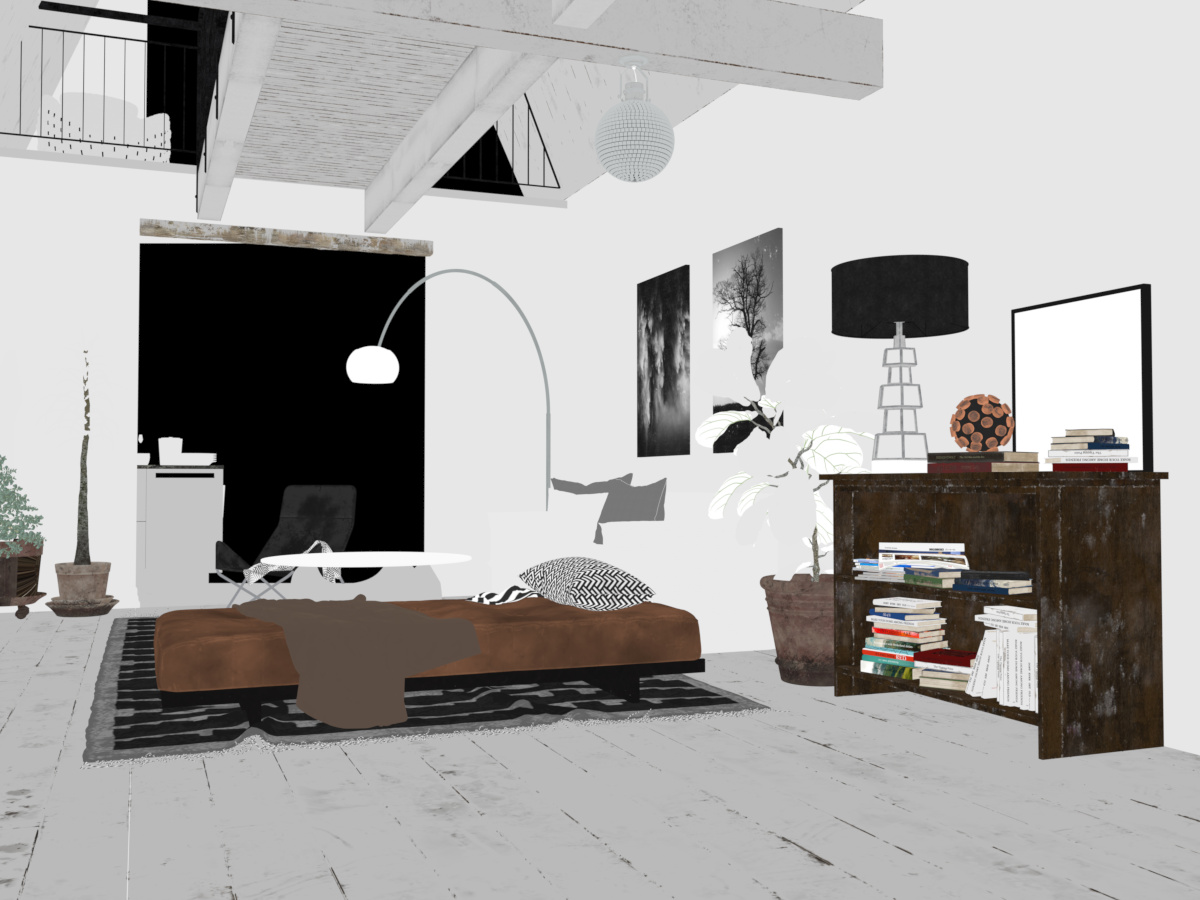}
};

\draw[signal] (531,419) -- (531,397);

\node[anchor=north west] at (1235.956,16.123) {%
  \includegraphics[width=1.183cm,height=0.887cm]
    {images/images_method/panel_04.jpg}
};
\node[anchor=north west] at (1231.847,20.170) {%
  \includegraphics[width=1.183cm,height=0.887cm]
    {images/images_method/panel_05.jpg}
};
\node[anchor=north west] at (1228.038,23.205) {%
  \includegraphics[width=1.183cm,height=0.887cm]
    {images/images_method/panel_06.jpg}
};
\node[anchor=north west] at (1224.519,27.251) {%
  \includegraphics[width=1.183cm,height=0.887cm]
    {images/images_method/panel_07.jpg}
};
\node[anchor=north west] at (1221.000,30.286) {%
  \includegraphics[width=1.183cm,height=0.887cm]
    {images/images_method/panel_08.jpg}
};

\node[
  text=mInk,
  font=\normalfont\fontsize{8.5}{9.5}\selectfont
] at (1268,3) {GTs};

\node[anchor=north west] at (1041.896,33.623) {%
  \includegraphics[width=1.080cm,height=0.810cm]
    {images/images_method/panel_04.jpg}
};
\node[anchor=north west] at (966.531,33.620) {%
  \includegraphics[width=1.080cm,height=0.810cm]
    {images/images_method/panel_05.jpg}
};
\node[anchor=north west] at (891.598,33.625) {%
  \includegraphics[width=1.080cm,height=0.810cm]
    {images/images_method/panel_06.jpg}
};
\node[anchor=north west] at (816.899,31.551) {%
  \includegraphics[width=1.080cm,height=0.810cm]
    {images/images_method/panel_07.jpg}
};
\node[anchor=north west] at (740.634,31.550) {%
  \includegraphics[width=1.080cm,height=0.810cm]
    {images/images_method/panel_08.jpg}
};

\draw[alignment] (610,233) -- (690,233);

\node[alignmenttext]
  at (650,198) {feature\\alignment};

\node[alignmenttext]
  (pixelLabel) at (1175.5,95.7) {pixel\\alignment};

\draw[alignment]
  (pixelLabel.west |- 0,60.7)
  --
  (pixelLabel.east |- 0,60.7);

\draw[draw=mNavy!45,->,line width=.75pt]
  (372,148) -- (861.000,148);

\draw[draw=mNavy!45,->,line width=.75pt]
  (369,366) -- (440,366);

\node[inputlabel]
  (rgbLabel) at (735,450) {RGB image};

\draw[signal] (rgbLabel.east) -- (836,450);

\node[inputlabel]
  (gtLabel) at (755,500) {GT Intrinsic Maps};

\draw[signal] (gtLabel.west) -- (635,500);

\node[
  align=center,
  text=mNavy,
  font=\normalfont\bfseries\fontsize{8.2}{9.6}\selectfont,
  text width=3.600cm
] at (301,588) {(a) $X\rightarrow X$ autoencoder};

\node[
  align=center,
  text=mNavy,
  font=\normalfont\bfseries\fontsize{8.2}{9.6}\selectfont,
  text width=11.550cm
] at (840,588) {(b) RGB $\rightarrow$ $X$ decomposition};

\path[projection]
  (720.600,260) rectangle (796.200,310);

\node[projectiontitle] at (758.400,277) {proj.};
\node[modality] at (758.400,299) {albedo};

\path[projection]
  (803.400,260) rectangle (879.000,310);

\node[projectiontitle] at (841.200,277) {proj.};
\node[modality] at (841.200,299) {irradiance};

\path[projection]
  (886.200,260) rectangle (961.800,310);

\node[projectiontitle] at (924.000,277) {proj.};
\node[modality] at (924.000,299) {normal};

\path[projection]
  (969.000,260) rectangle (1044.600,310);

\node[projectiontitle] at (1006.800,277) {proj.};
\node[modality] at (1006.800,299) {roughness};

\path[projection]
  (1051.800,260) rectangle (1127.400,310);

\node[projectiontitle] at (1089.600,277) {proj.};
\node[modality] at (1089.600,299) {metallic};

\node[anchor=north west] at (317.422,430.042) {%
  \includegraphics[width=0.858cm,height=0.644cm]
    {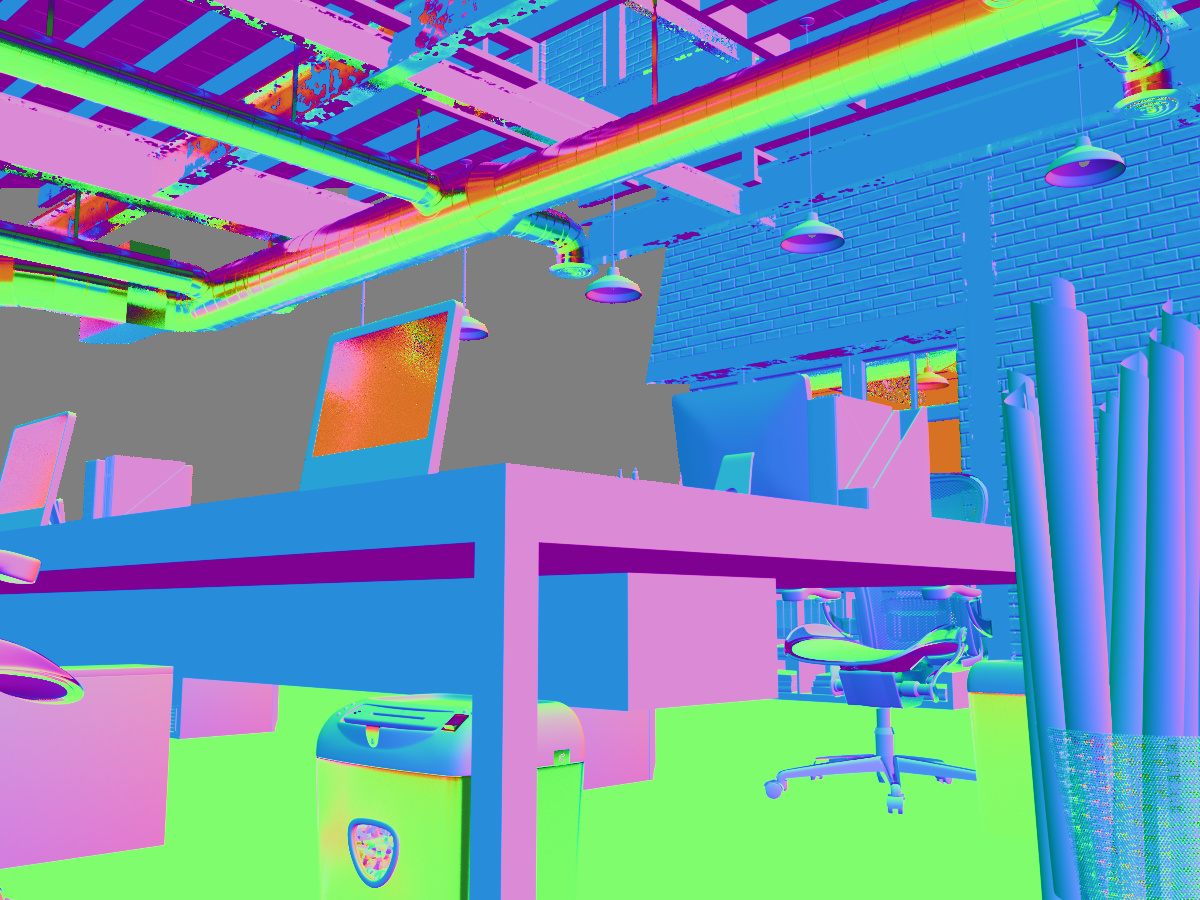}
};
\node[anchor=north west] at (225.910,493.940) {%
  \includegraphics[width=0.858cm,height=0.644cm]
    {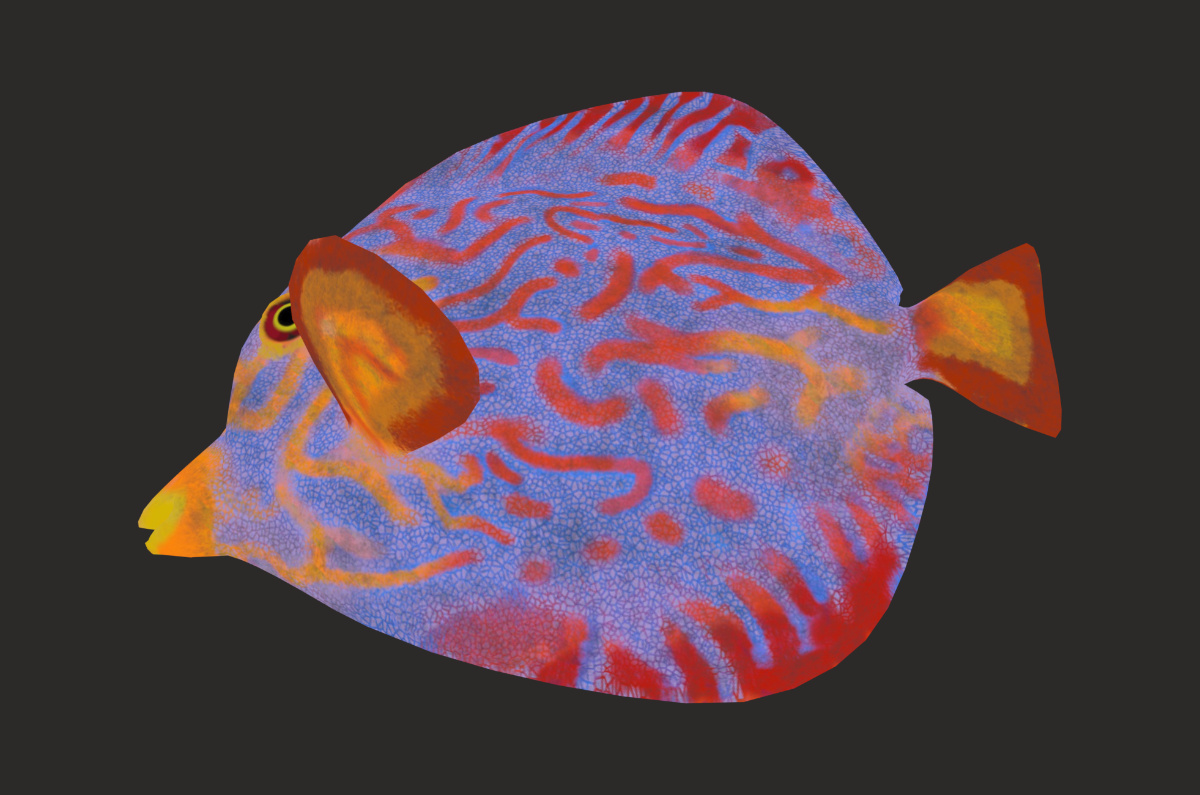}
};

\draw[signal] (302,419) -- (302,397);

\node[anchor=north west] at (225.910,432.107) {%
  \includegraphics[width=0.858cm,height=0.644cm]
    {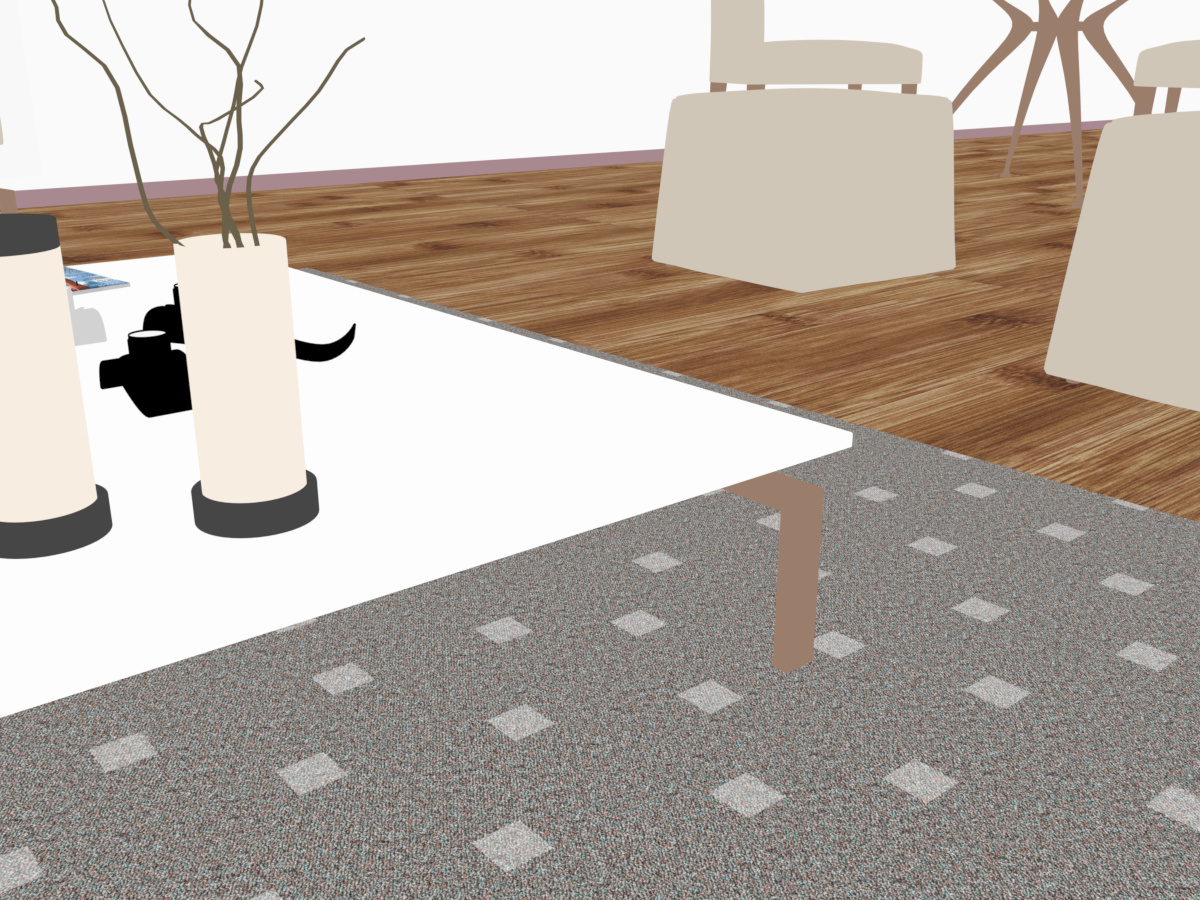}
};
\node[anchor=north west] at (326.706,455.176) {%
  \includegraphics[width=0.858cm,height=0.644cm]
    {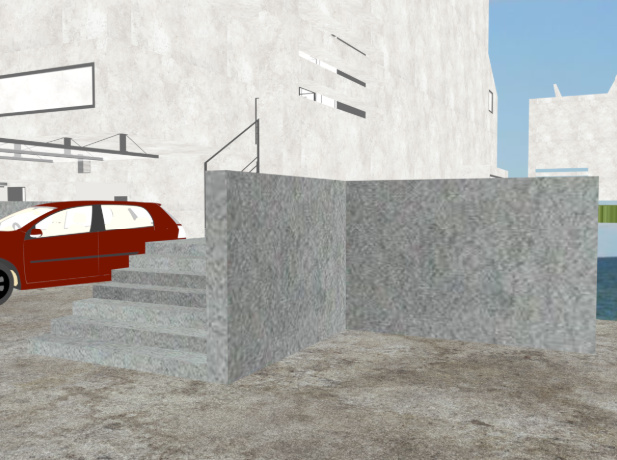}
};
\node[anchor=north west] at (258.555,500.501) {%
  \includegraphics[width=0.858cm,height=0.644cm]
    {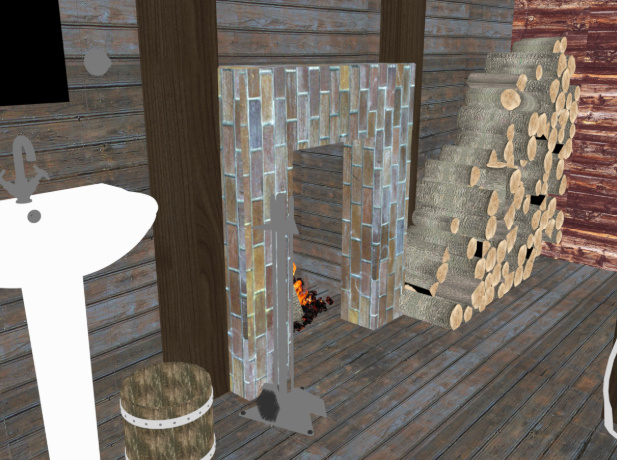}
};
\node[anchor=north west] at (242.642,455.178) {%
  \includegraphics[width=0.858cm,height=0.644cm]
    {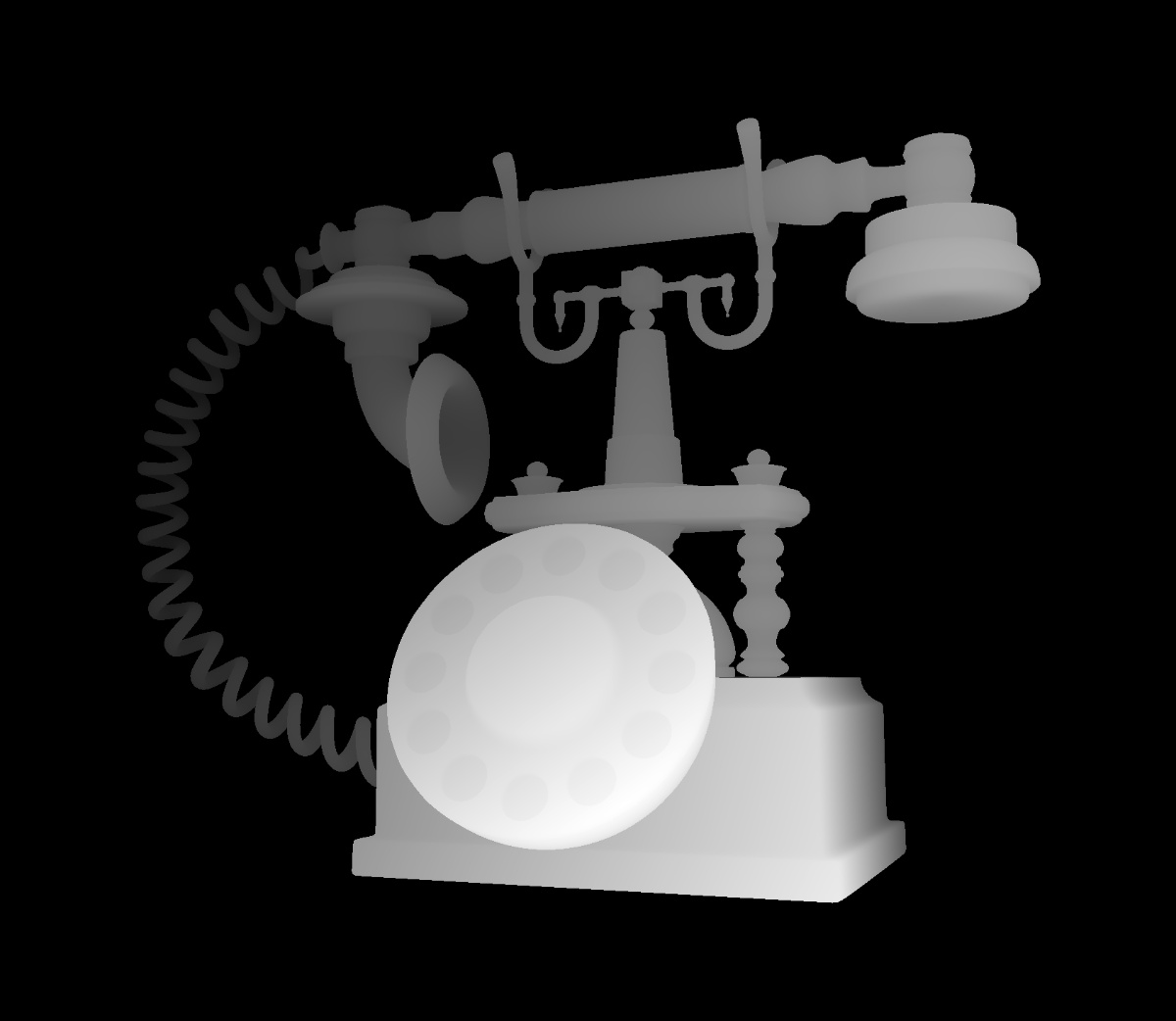}
};
\node[anchor=north west] at (320.830,485.874) {%
  \includegraphics[width=0.858cm,height=0.644cm]
    {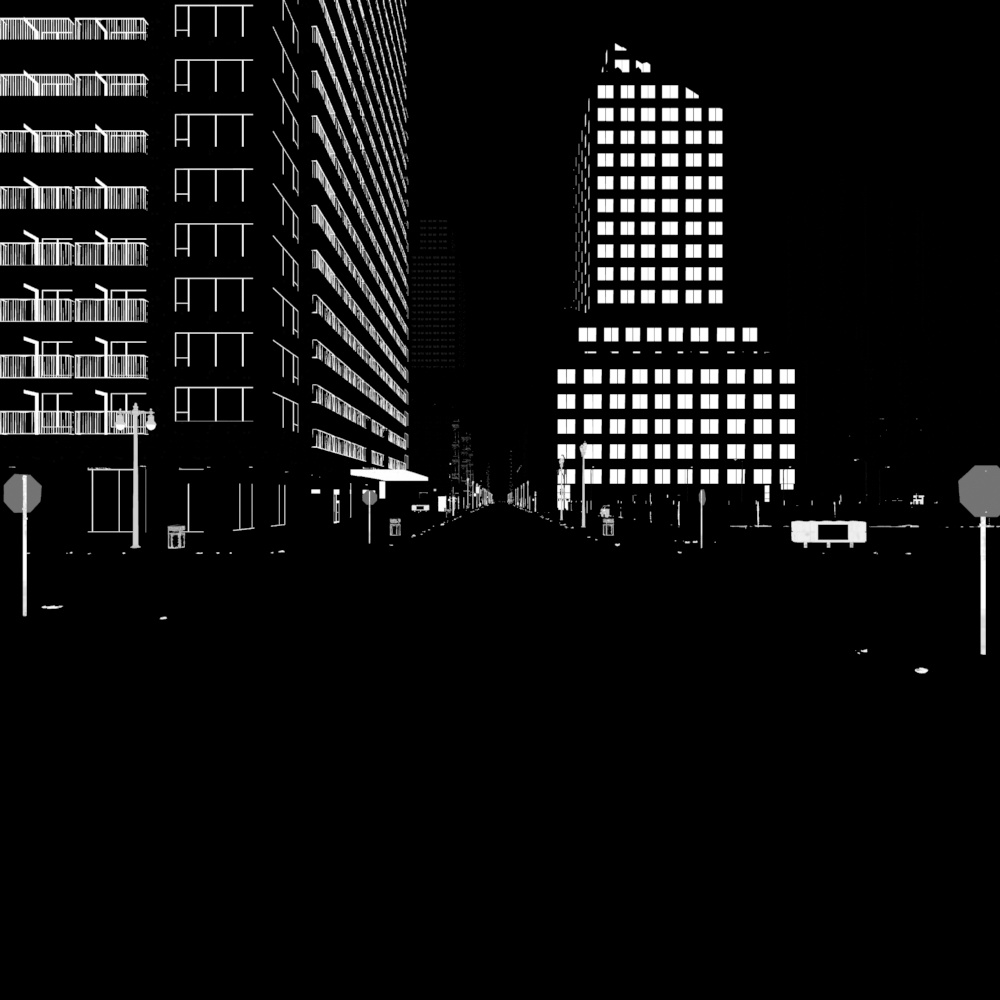}
};
\node[anchor=north west] at (270.297,472.946) {%
  \includegraphics[width=0.858cm,height=0.644cm]
    {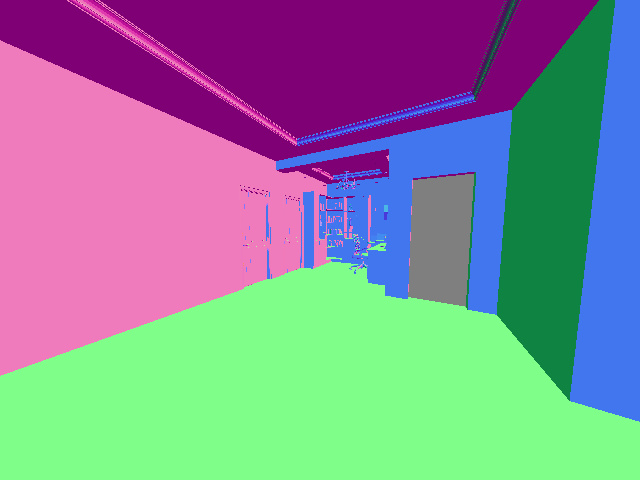}
};
\node[anchor=north west] at (305.401,510.096) {%
  \includegraphics[width=0.858cm,height=0.644cm]
    {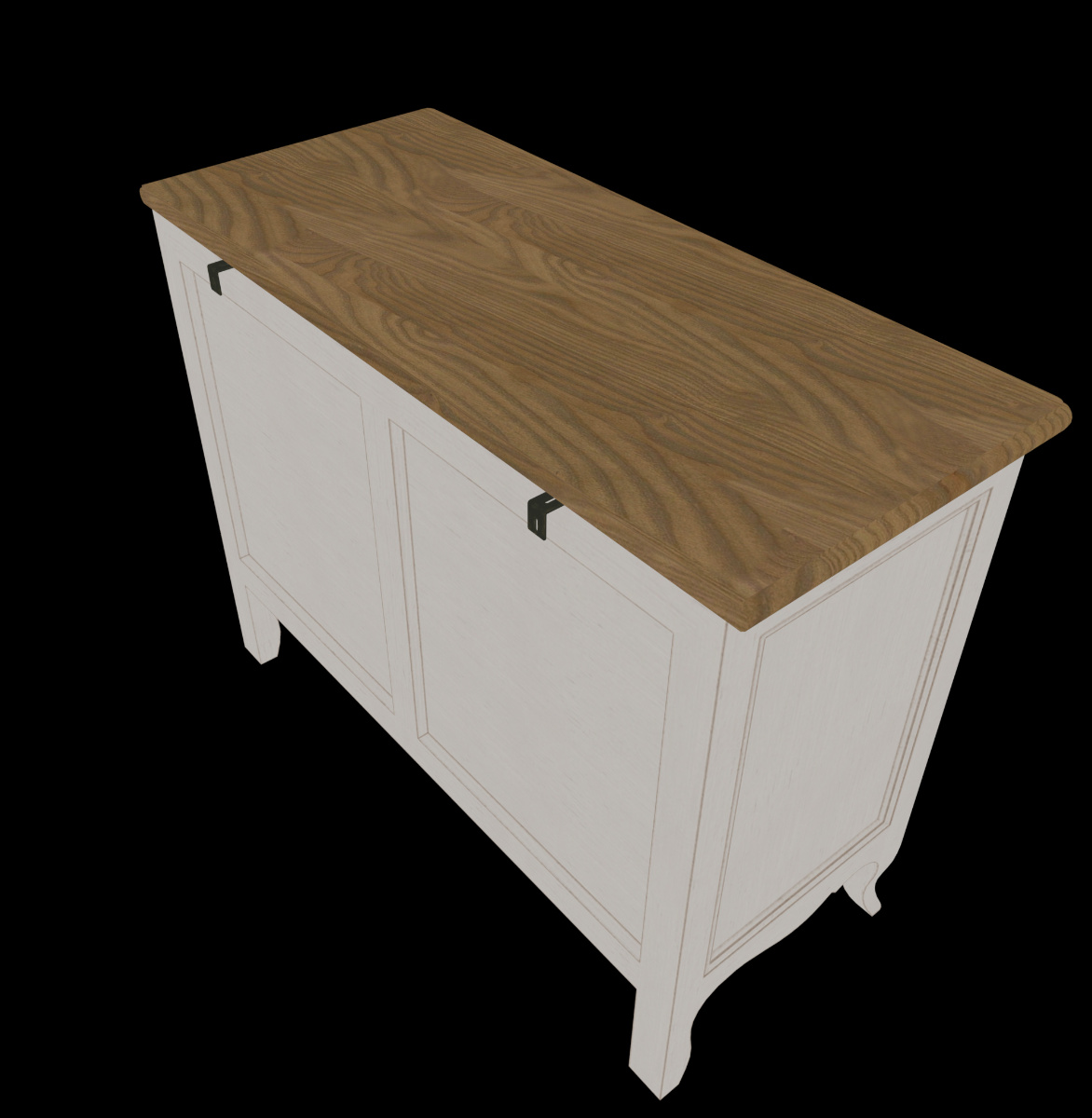}
};
\node[anchor=north west] at (265.297,419.870) {%
  \includegraphics[width=1.131cm,height=0.848cm]
    {images/images_method/panel_02.jpg}
};

\draw[draw=mInk,line width=.7pt]
  (758.400,329) -- (1089.600,329);

\draw[draw=mInk,line width=.72pt]
  (924.000,337) -- (924.000,329);

\draw[signal] (758.400,329)  -- (758.400,311);
\draw[signal] (841.200,329)  -- (841.200,311);
\draw[signal] (924.000,329)  -- (924.000,311);
\draw[signal] (1006.800,329) -- (1006.800,311);
\draw[signal] (1089.600,329) -- (1089.600,311);

\draw[signal] (776.400,114.840)  -- (776.400,94.000);
\draw[signal] (852.900,115.530)  -- (852.900,94.000);
\draw[signal] (927.033,112.840)  -- (927.033,94.000);
\draw[signal] (1002.012,113.000) -- (1002.012,94.000);
\draw[signal] (1077.612,113.690) -- (1077.612,94.000);

\node[anchor=north west] at (514.517,152.953) {%
  \includegraphics[width=1.256cm,height=2.344cm]
    {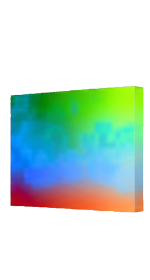}
};
\node[anchor=north west] at (492.553,153.377) {%
  \includegraphics[width=1.256cm,height=2.344cm]
    {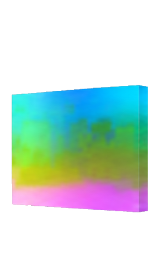}
};
\node[anchor=north west] at (473.544,153.377) {%
  \includegraphics[width=1.256cm,height=2.344cm]
    {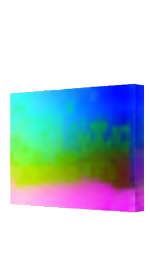}
};
\node[anchor=north west] at (456.115,152.530) {%
  \includegraphics[width=1.256cm,height=2.344cm]
    {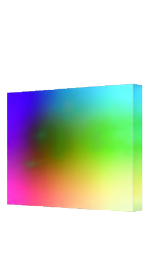}
};
\node[anchor=north west] at (438.000,152.953) {%
  \includegraphics[width=1.256cm,height=2.344cm]
    {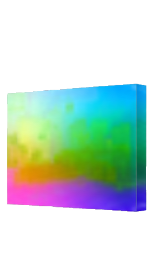}
};

\node[anchor=north west] at (1056.378,150.398) {%
  \includegraphics[width=1.124cm,height=2.098cm]
    {images/images_method/panel_19.png}
};
\node[anchor=north west] at (974.474,149.996) {%
  \includegraphics[width=1.124cm,height=2.098cm]
    {images/images_method/panel_20.png}
};
\node[anchor=north west] at (893.948,152.526) {%
  \includegraphics[width=1.124cm,height=2.098cm]
    {images/images_method/panel_21.png}
};
\node[anchor=north west] at (807.901,150.000) {%
  \includegraphics[width=1.124cm,height=2.098cm]
    {images/images_method/panel_22.png}
};
\node[anchor=north west] at (722.400,150.398) {%
  \includegraphics[width=1.124cm,height=2.098cm]
    {images/images_method/panel_23.png}
};

\path[decoder] (1190,525) rectangle (1218,545);
\node[anchor=west] at (1222,535) {\methodFlame};
\node[inputlabel,anchor=west] at (1247,535) {trainable};
\path[model] (1190,561) rectangle (1218,581);
\methodSnowflake{1234.5}{571}
\node[inputlabel,anchor=west] at (1247,571) {frozen};

\end{tikzpicture}%
}

\endgroup
\caption{\textbf{\methodName\ overview.}
(a) We first learn an $X\!\rightarrow\!X$ reconstruction autoencoder across RGB and intrinsic signals, \ie albedo, irradiance, normals, roughness, and metallicity. A frozen DINOv3 backbone with trainable AdaptFormer modules encodes each modality into a shared, decodable feature space; one DC-AE decoder reconstructs the input modality from that space.
(b) We then freeze the reconstruction model and learn to infer its features from RGB. A student encoder and modality-specific projections predict the features that the teacher extracts from the corresponding GT maps. Feature alignment guides these predictions, while supervision on the decoded maps enforces pixel alignment. At inference, the teacher is removed: the student, projections, and same shared decoder produce all five maps.}
\label{fig:method}
\vspace{-0.8em}
\end{figure*}

\myparagraph{Intrinsic scene understanding.}
Intrinsic decomposition aims to recover physical scene factors from an RGB image, classically separating reflectance from illumination~\cite{Barrow:1978:Intrinsic, Land:1971:Retinex}. Early methods relied on hand-designed priors, while learning-based approaches recast the problem as dense prediction from paired or weakly supervised data~\cite{Li:2018:CGIntrinsics, Li:2020:InverseRenderingIndoor, Li:2021:OpenRooms, Roberts:2021:Hypersim, Zhu:2022:IRISFormer, Zhu:2022:InteriorVerse, Careaga:2023:OrdinalShading, Careaga:2024:ColorfulIID}. Recent works extend the decomposition beyond albedo and shading to geometry, illumination, and material properties, including normals, irradiance, roughness, metallicity, and depth~\cite{Dirik:2026:PRISM, Zeng:2024:rgbx, wu:2025:MVInverse, dirik:2026:geoid}. A prominent recent direction is to adapt generative foundation models to inverse rendering. Diffusion-based methods repurpose pretrained image generators for depth, normals, albedo, shading, and PBR-related maps~\cite{Ke:2024:MarigoldDepth, Ye:2024:StableNormal, Kocsis:2024:IntrinsicImageDiffusion, Luo:2024:IntrinsicDiffusion, Zeng:2024:rgbx, Dirik:2026:PRISM}. These models benefit from strong generative priors, which help produce plausible intrinsic maps. However, decomposition typically remains an iterative sampling procedure, making inference stochastic, computationally expensive, and difficult to scale to high resolution. Single-step variants~\cite{He:2025:Lotus, Pavlovic:2026:MarigoldV2} reduce sampling cost but retain large generative backbones; the concurrent Marigold V2~\cite{Pavlovic:2026:MarigoldV2} is two orders of magnitude larger than our model. Feed-forward inverse rendering has also been explored: IRISformer~\cite{Zhu:2022:IRISFormer} jointly predicts geometry, materials, and lighting from a single indoor image, while MVInverse~\cite{wu:2025:MVInverse} targets multi-view inputs with a large-scale transformer. Our work addresses the single-image setting with a compact feed-forward model, predicting appearance, geometry, illumination, and material properties jointly from DINOv3 features, without generative sampling.

\myparagraph{Vision foundation models.}
The recent progress of vision foundation models has shown that large-scale pretraining provides strong priors for a wide range of downstream tasks. One line of work builds on generative models, particularly latent diffusion models~\cite{Rombach:2022:LDM}, whose synthesis priors help resolve ambiguities in dense inverse problems, as discussed above, at the cost of iterative and stochastic inference.
A complementary line of work learns general-purpose visual representations through large-scale self-supervised or multimodal pretraining, achieving strong transfer across recognition and dense prediction tasks~\cite{Caron:2021:DINO, He:2022:MAE, Zhou:2021:iBOT, Oquab:2023:Dinov2, Tschannen:2025:SigLIP2, Heinrich:2025:RADIOv25}. 
Among these, DINOv3~\cite{Simeoni:2025:Dinov3} combines high-level semantic abstraction with dense, localized features, supporting reasoning about objects and materials while retaining pixel-aligned spatial structure. DINO features have also been used as reconstruction latents~\cite{Zheng:2026:RAE} and alignment targets~\cite{Yu:2025:REPA}, which our teacher builds on.

\section{Method}
\label{sec:method}

\begin{figure}[t]
    \centering
    \includegraphics[width=\linewidth]{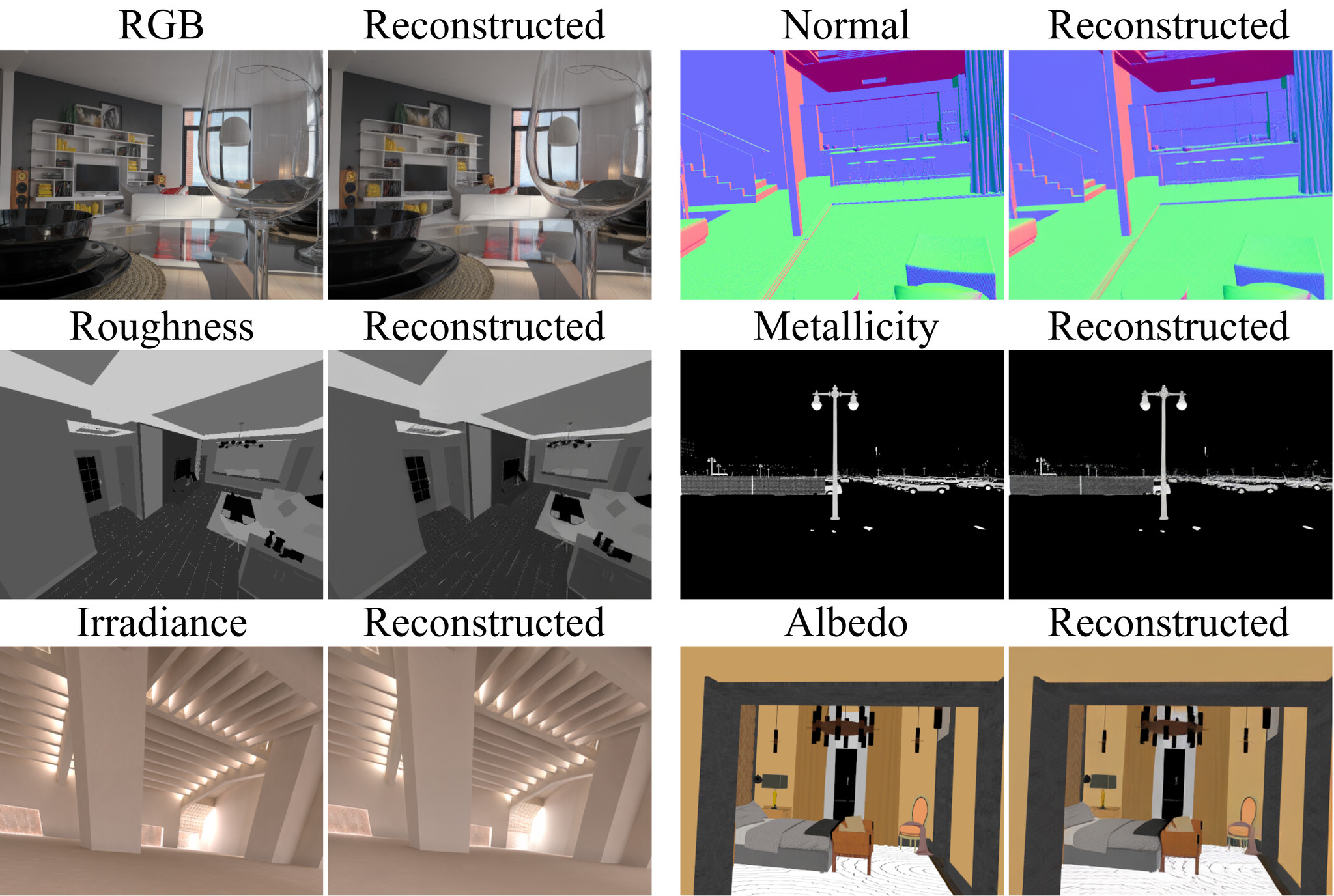}
    \caption{\textbf{\boldmath $X \rightarrow X$ autoencoder reconstructions.}
    The reconstruction stage receives a map $X$ as input and reconstructs the same signal at the output.
    The autoencoder preserves the spatial structure and modality-specific details of the input maps. These reconstructions indicate that heterogeneous intrinsic signals can be represented in a shared DINOv3 feature-to-map space, which is later used to supervise the RGB-to-intrinsics student.}

\label{fig:reconstruction_autoencoder}
\end{figure}

\myparagraph{Overview.}
Our model, \methodName, is summarized in \cref{fig:method}. Given a single RGB image, it predicts albedo, surface normals, irradiance, roughness, and metallicity in a single feed-forward pass.
The key idea is to decouple the inference of intrinsic properties from their reconstruction into pixel-space maps. 
We first train an $X \rightarrow X$ autoencoder built on DINOv3, pairing its pretrained encoder with a shared DC-AE decoder~\cite{Chen:2025:DC-AE}. Given an input modality, the encoder extracts dense feature maps, from which the decoder reconstructs the original map (\cref{fig:reconstruction_autoencoder}). Training the autoencoder across modalities establishes a common, decodable feature space, without requiring paired $\mathrm{RGB} \rightarrow X$ supervision.
We then freeze the reconstruction decoder and train the RGB student. From the DINOv3 representation of the input image, modality-specific projection heads predict the teacher features extracted from each ground-truth intrinsic map. We supervise the student to predict the modality-specific representations from the RGB features, so that the frozen reconstruction decoder can map the predicted features into their corresponding intrinsic maps.
Both teacher and student use a frozen DINOv3 with lightweight adapters in the final layers, preserving the pretrained structure while specializing the representation to intrinsic prediction. The student is trained at variable resolution to encourage stable feature alignment across different spatial grids.
At inference time, the teacher encoder is discarded. The final model consists of the RGB encoder, the modality-specific projection heads, and the decoder learned in the reconstruction stage, which is applied to every output modality.

\myparagraph{Task formulation.}
Given an RGB image $I \in [0,1]^{3 \times H \times W}$, our goal is to recover a set of physically meaningful intrinsic maps. We denote the set of target modalities as
\[
\begin{aligned}
\mathcal{M} = \{&
\mathrm{albedo}, \mathrm{normal}, \mathrm{irradiance}, \mathrm{roughness}, \mathrm{metallic}
\}.
\end{aligned}
\]
For each modality $m \in \mathcal{M}$, let $X^m$ be the corresponding ground-truth map and $\hat{X}^m$ its prediction. The decomposition is a deterministic mapping
\[
F(I) = \{\hat{X}^m\}_{m \in \mathcal{M}}.
\]

\myparagraph{\boldmath $X \rightarrow X$ autoencoder for reconstruction.}
The purpose of the first stage is to learn a reusable reconstruction space for dense map-like signals. We train a teacher autoencoder that receives a target map as input and reconstructs the same map at the output. The autoencoder is trained on the modality set $\mathcal{M} \cup \{\mathrm{RGB}\}$.
Because the target modalities have different physical domains, we first map each of them to a common canonical image space. For each modality $m \in \mathcal{M}$, we define a canonicalization function $c_m(\cdot)$ and write
\begin{equation}
\tilde{X}^m = c_m(X^m) \in [0,1]^{3 \times H \times W}.
\end{equation}
Albedo and irradiance are represented as sRGB-encoded images, normals are encoded in the $[0,1]$ range, and scalar maps such as roughness and metallicity are converted to three-channel image-like fields. This common representation allows a single reconstruction model to be trained across appearance, geometry, illumination, and material signals. The reconstruction teacher is defined as
\begin{equation}
T = (E_T, D),
\end{equation}
where $E_T$ is a DINOv3 encoder and $D$ is a DC-AE decoder~\cite{Chen:2025:DC-AE}. The DC-AE decoder efficiently reconstructs high-resolution images from compact latent feature maps, making it suitable for dense intrinsic-map reconstruction.
We keep the pretrained DINOv3 backbone frozen and train only lightweight AdaptFormer modules~\cite{Chen:2022:AdaptFormer} in its final transformer layers. 

Given a canonicalized map $\tilde{X}^m$, the teacher extracts a dense feature field and reconstructs the input:
\begin{align}
Z_T^m &= E_T(\tilde{X}^m), \\
\hat{\tilde{X}}^m &= \sigma(D(Z_T^m)),
\end{align}
where $\sigma$ maps the decoder output to the $[0,1]$ range. \cref{fig:reconstruction_autoencoder} shows qualitative reconstructions produced by the trained autoencoder across the different input modalities.

The reconstruction objective combines pixel and perceptual terms:
\begin{equation}
\mathcal{L}_{X \rightarrow X}
=
\mathcal{L}_{1}
\left(
\hat{\tilde{X}}^m,
\tilde{X}^m
\right)
+
\mathcal{L}_{\mathrm{LPIPS}}
\left(
\hat{\tilde{X}}^m,
\tilde{X}^m
\right).
\end{equation}

After this stage, the decoder $D$ acts as a shared feature-to-map renderer, and the teacher encoder $E_T$ defines the modality-specific target feature fields.

\myparagraph{\boldmath $\mathrm{RGB} \rightarrow X$ decomposition.}
The second stage learns the inverse mapping from an RGB image to the intrinsic maps. Each training example contains an RGB observation $I$ and the maps $\{X^m\}$ available in its source dataset; losses are applied only to the available modalities, so partially annotated datasets can be combined. 
The reconstruction teacher from the previous stage is kept fixed and provides for each target map, a feature representation that can be decoded into pixel space.
The student encoder $E_S$ shares the architecture of the teacher $E_T$, including the adapters, and is trained to predict the teacher features from RGB.

Given an image $I$, the student encoder extracts features
\begin{equation}
Z_S = E_S(I).
\end{equation}
This representation is shared across all target modalities. To obtain modality-specific predictions, we introduce one projection head $P_m$ for each map $m \in \mathcal{M}$. Each projection maps the shared RGB feature field into the reconstruction feature space of the corresponding modality.

We implement each projection $P_m$ as a stack of $K$ lightweight residual convolutional blocks. Starting from $Y_0^m = Z_S$, the projection is defined as
\begin{equation}
Y_k^m =
Y_{k-1}^m
+
R_k^m(Y_{k-1}^m),
\qquad
k = 1,\dots,K,
\end{equation}
where $R_k^m$ denotes the residual transformation of the $k$-th block. The final modality-specific feature field is
\begin{equation}
\hat{Z}^m = P_m(Z_S) = Y_K^m.
\end{equation}
Since each modality has its own projection, the shared RGB representation is decomposed into distinct feature fields for albedo, normal, irradiance, roughness, and metallicity. The projected feature is decoded by the frozen reconstruction decoder:
\begin{align}
\hat{\tilde{X}}^m &= \sigma(D(\hat{Z}^m)), \\
\hat{X}^m &= c_m^{-1}(\hat{\tilde{X}}^m).
\end{align}

The target feature for each modality is obtained by encoding the canonicalized ground-truth map with the frozen teacher encoder, $Z_T^m = E_T(c_m(X^m))$.
It represents the point in the reconstruction space that corresponds to the true intrinsic map, and the student is trained to predict it from RGB, optimizing only its adapters and the projection heads. 

We supervise the student in both feature space and output space. The feature-space objective aligns the projected student features with the teacher features:
\begin{equation}
\mathcal{L}_{\mathrm{feat}}^m
=
\mathcal{L}_{\mathrm{MSE}}(\hat{Z}^m, Z_T^m)
+
\mathcal{L}_{\mathrm{cos}}(\hat{Z}^m, Z_T^m).
\end{equation}
The MSE term matches the teacher representation locally, while the cosine term encourages directional agreement in the reconstruction feature space.

In addition to feature alignment, we apply direct supervision on the decoded maps:
\begin{equation}
\mathcal{L}_{\mathrm{out}}^m
=
\mathcal{D}_m(\hat{X}^m, X^m)
+
\mathcal{G}_m(\hat{X}^m, X^m)
+
\mathcal{L}_{\mathrm{LPIPS}}(\hat{X}^m, X^m).
\end{equation}
The data term $\mathcal{D}_m$ is an $\ell_1$ loss for albedo, irradiance, roughness, and metallicity, and an angular error for surface normals. The term $\mathcal{G}_m$ is a multi-scale gradient loss~\cite{wu:2025:MVInverse, Careaga:2024:ColorfulIID, Li:2018:CGIntrinsics, Li:2018:MegaDepth}, used to preserve spatial discontinuities and fine structures.

The final student objective for modality $m$ is
\begin{equation}
\mathcal{L}_{\mathrm{RGB} \rightarrow X}^m
=
\mathcal{L}_{\mathrm{feat}}^m
+
\mathcal{L}_{\mathrm{out}}^m.
\end{equation}

We train the student with multi-resolution supervision, sampling the input size per iteration, which encourages features that remain stable across spatial grids.

\myparagraph{Inference.}
At inference time, the teacher is discarded. The final model consists of the adapted student backbone $E_S$, the map-specific projection heads $\{P_m\}_{m \in \mathcal{M}}$, and the pretrained reconstruction decoder $D$. Given an RGB image, a single encoder pass followed by one decoder evaluation per modality produces all intrinsic maps:

\begin{equation}
\begin{gathered}
Z_S = E_S(I), \\
\hat{X}^m =
c_m^{-1}\!\left(\sigma\!\left(D\!\left(P_m(Z_S)\right)\right)\right),
\qquad m \in \mathcal{M}.
\end{gathered}
\end{equation}
\begin{table*}[t]
\centering
\caption{\textbf{Comparison on InteriorVerse and Hypersim.}
\methodName{} offers the best overall accuracy--efficiency trade-off, maintaining competitive albedo and normal estimation while achieving the best material estimates (roughness and metallicity) on InteriorVerse and irradiance estimates on Hypersim.
With only $165$M parameters, it is the smallest and fastest model in this comparison, delivering over $100{\times}$ the throughput of the fastest diffusion baseline~\cite{Zeng:2024:rgbx} on InteriorVerse.
Compared with the feed-forward baseline~\cite{wu:2025:MVInverse}, it uses about one-sixth as many parameters and achieves $2{\times}$ FPS at $1024{\times}768$.
For albedo and irradiance, we report PSNR, RMSE, and LPIPS; for metallicity and roughness, PSNR, RMSE, and MAE; and for normals, mean angular error and pixel accuracy at $11.25^\circ$ and $30^\circ$.
FPS is measured on an NVIDIA A100 80GB GPU at $640{\times}480$ on InteriorVerse and $1024{\times}768$ on Hypersim.
Best results \textbf{bold}, \nth{2} best \underline{underlined}.}
\label{tab:intrinsic_quantitative}
\label{tab:interiorverse_quantitative}
\label{tab:hypersim_quantitative}
\vspace{-0.5em}
\small
\setlength{\tabcolsep}{3.0pt}
\resizebox{\linewidth}{!}{%
\begin{tabular}{@{}lcc|ccc|ccc|ccc|ccc@{}}
\toprule
\rowcolor{black!8}[0pt][0pt]
\multicolumn{15}{@{}l@{}}{\textbf{InteriorVerse} \quad ($640{\times}480$)} \\
\midrule
& & &
\multicolumn{3}{c|}{\textbf{Albedo}} &
\multicolumn{3}{c|}{\textbf{Normal}} &
\multicolumn{3}{c|}{\textbf{Metallicity}} &
\multicolumn{3}{c}{\textbf{Roughness}} \\
\cmidrule(lr){4-6}\cmidrule(lr){7-9}\cmidrule(lr){10-12}\cmidrule(lr){13-15}
\textbf{Method} & \textbf{\#Param} $\downarrow$ & \textbf{FPS} $\uparrow$ &
PSNR $\uparrow$ & RMSE $\downarrow$ & LPIPS $\downarrow$ &
Mean $\downarrow$ & $11.25^\circ$ $\uparrow$ & $30^\circ$ $\uparrow$ &
PSNR $\uparrow$ & RMSE $\downarrow$ & MAE $\downarrow$ &
PSNR $\uparrow$ & RMSE $\downarrow$ & MAE $\downarrow$ \\
\midrule
ColorfulDiffuse~\cite{Careaga:2024:ColorfulIID}

& 443M & 0.90
& 12.29 & 0.247 & 0.209
& -- & -- & --
& -- & -- & --
& -- & -- & -- \\

IntrinsicImageDiffusion~\cite{Kocsis:2024:IntrinsicImageDiffusion}

& 1.7B & 0.03
& 13.26 & 0.229 & 0.196
& -- & -- & --
& 16.47 & 0.158 & 0.126
& 13.15 & 0.231 & 0.193 \\

DiffusionRenderer \cite{Liang:2025:DiffusionRenderer}

& 1.6B & 0.10
& \textbf{21.19} & \textbf{0.094} & 0.106
& \underline{9.11} & 82.57 & \textbf{93.28}
& 10.57 & 0.361 & 0.285
& 11.13 & 0.287 & 0.248 \\

RGB$\leftrightarrow$X~\cite{Zeng:2024:rgbx}

& 1.3B & 0.13
& 19.74 & 0.113 & \underline{0.099}
& 9.14 & \textbf{84.15} & \underline{92.74}
& 6.08 & 0.501 & 0.462
& 10.94 & 0.294 & 0.252 \\

MVInverse~\cite{wu:2025:MVInverse}

& 1B & \underline{8.96}
& 17.12 & 0.148 & 0.172
& 12.02 & 71.67 & 90.24
& \underline{18.75} & \underline{0.123} & \underline{0.079}
& \underline{15.77} & \underline{0.172} & \underline{0.136} \\

\midrule

\textbf{\methodName}

& \textbf{165M} & \textbf{14.97}
& \underline{20.92} & \underline{0.095} & \textbf{0.092}
& \textbf{8.55} & \underline{83.14} & 92.06
& \textbf{20.82} & \textbf{0.099} & \textbf{0.045}
& \textbf{19.67} & \textbf{0.114} & \textbf{0.061} \\
\midrule
\rowcolor{black!8}[0pt][0pt]
\multicolumn{15}{@{}l@{}}{\textbf{Hypersim} \quad ($1024{\times}768$)} \\
\midrule
& & &
\multicolumn{3}{c|}{\textbf{Albedo}} &
\multicolumn{3}{c|}{\textbf{Normal}} &
\multicolumn{6}{c}{\textbf{Irradiance}} \\
\cmidrule(lr){4-6}\cmidrule(lr){7-9}\cmidrule(lr){10-15}
\textbf{Method} & \textbf{\#Param} $\downarrow$ & \textbf{FPS} $\uparrow$ &
PSNR $\uparrow$ & RMSE $\downarrow$ & LPIPS $\downarrow$ &
Mean $\downarrow$ & $11.25^\circ$ $\uparrow$ & $30^\circ$ $\uparrow$ &
\multicolumn{2}{c}{PSNR $\uparrow$} & \multicolumn{2}{c}{RMSE $\downarrow$} & \multicolumn{2}{c}{LPIPS $\downarrow$} \\
\midrule
ColorfulDiffuse~\cite{Careaga:2024:ColorfulIID}

& 443M & 0.83
& 15.86 & 0.173 & 0.317
& -- & -- & --
& \multicolumn{2}{c}{12.18} & \multicolumn{2}{c}{0.295} & \multicolumn{2}{c}{0.335} \\

IntrinsicImageDiffusion~\cite{Kocsis:2024:IntrinsicImageDiffusion}

& 1.7B & 0.03
& 11.55 & 0.278 & 0.451
& -- & -- & --
& \multicolumn{2}{c}{--} & \multicolumn{2}{c}{--} & \multicolumn{2}{c}{--} \\

DiffusionRenderer \cite{Liang:2025:DiffusionRenderer}

& 1.6B & 0.10
& 12.95 & 0.241 & 0.375
& 17.61 & \underline{68.18} & 82.87
& \multicolumn{2}{c}{--} & \multicolumn{2}{c}{--} & \multicolumn{2}{c}{--} \\

MVInverse~\cite{wu:2025:MVInverse}

& 1B & \underline{4.45}
& \textbf{16.58} & \underline{0.168} & \underline{0.253}
& \textbf{12.44} & \textbf{74.56} & \textbf{88.10}
& \multicolumn{2}{c}{\underline{15.43}} & \multicolumn{2}{c}{\underline{0.182}} & \multicolumn{2}{c}{\underline{0.321}} \\

\midrule

\textbf{\methodName}

& \textbf{165M} & \textbf{8.94}
& \underline{16.32} & \textbf{0.163} & \textbf{0.211}
& \underline{16.90} & 66.79 & \underline{83.65}
& \multicolumn{2}{c}{\textbf{17.86}} & \multicolumn{2}{c}{\textbf{0.149}} & \multicolumn{2}{c}{\textbf{0.231}} \\
\bottomrule
\end{tabular}%
}
\vspace{0.2em}
\end{table*}

\section{Implementation \& Results}

\myparagraph{Implementation details.}
We use DINOv3-Base~\cite{Simeoni:2025:Dinov3} as the backbone in both stages and keep the pretrained backbone frozen ($86$M parameters). AdaptFormer modules~\cite{Chen:2022:AdaptFormer} are inserted in the final six transformer layers. The reconstruction stage trains the adapters ($3.6$M parameters) and the DC-AE decoder~\cite{Chen:2025:DC-AE} ($28.7$M); the decomposition stage freezes the teacher and trains new adapters ($3.6$M) and the projection heads ($47.6$M). Both stages use AdamW (batch size $16$, bf16) on NVIDIA A100 GPUs: the reconstruction stage runs for $112.5$k iterations at learning rate $10^{-3}$ and a long side of $512$ pixels, and the decomposition stage for $281$k iterations at $10^{-4}$ with a long side sampled per iteration from $\{512, 1024\}$. Full training details are in the supplement.

\myparagraph{Datasets.}
We train on $2.3$M synthetic images spanning indoor scenes, object-centric assets, and urban environments, from Hypersim~\cite{Roberts:2021:Hypersim}, InteriorVerse~\cite{Zhu:2022:InteriorVerse}, Evermotion~\cite{Zeng:2024:rgbx}, an $88.7$k-image subsample of Objaverse~\cite{deitke:2022:objaverse}, ABO~\cite{Collins:2022:ABO}, and MatrixCity~\cite{Li:2023:MatrixCity}. Only Objaverse provides all five maps; the other datasets are partially annotated (see supplementary material), and each image is supervised only on its available maps.

\begin{figure}[t]
\centering
\begingroup

\definecolor{resolutionblue}{HTML}{337EAA}
\definecolor{resolutionorange}{HTML}{D48036}
\definecolor{resolutionours}{HTML}{BD4055}
\definecolor{resolutionpurple}{HTML}{8973AD}
\definecolor{resolutionink}{HTML}{293B4C}

\pgfplotstableread[col sep=comma,row sep=\\]{
index,resolution,mvinverse_psnr,mvinverse_lpips,mvinverse_memory,ours_psnr,ours_lpips,ours_memory,single_psnr,single_lpips,single_memory\\
0,512,15.013,0.25504,6.498,16.128,0.21029,1.464,16.235,0.21466,1.464\\
1,640,15.281,0.24672,6.723,16.254,0.20742,1.520,16.403,0.21392,1.520\\
2,768,15.383,0.23252,6.989,16.383,0.20333,1.586,16.510,0.21201,1.586\\
3,1024,15.062,0.25911,7.697,16.489,0.20571,1.757,16.384,0.22057,1.757\\
4,1280,14.666,0.24558,8.606,16.480,0.20771,1.976,16.478,0.23745,1.976\\
5,1536,13.229,0.30418,9.707,16.395,0.21530,2.242,16.227,0.26138,2.242\\
6,2048,11.715,0.38548,12.520,15.756,0.23953,2.920,14.895,0.33009,2.920\\
}\resolutionhypersim

\pgfplotstableread[col sep=comma,row sep=\\]{
index,resolution,mvinverse_psnr,mvinverse_lpips,mvinverse_memory,rgbx_psnr,rgbx_lpips,rgbx_memory,ours_psnr,ours_lpips,ours_memory,single_psnr,single_lpips,single_memory\\
0,512,16.315,0.17313,6.498,18.935,0.11148,3.092,20.994,0.08732,1.464,20.122,0.09971,1.464\\
1,640,16.429,0.17392,6.723,19.489,0.09946,3.392,21.166,0.08950,1.520,20.282,0.09664,1.520\\
2,768,16.444,0.15894,6.989,17.769,0.10218,3.703,21.912,0.07695,1.586,20.782,0.09897,1.586\\
3,1024,16.313,0.16127,7.697,15.255,0.11566,4.580,22.292,0.07424,1.757,20.922,0.10855,1.757\\
4,1280,16.078,0.16966,8.606,14.634,0.13108,5.684,22.427,0.07764,1.976,20.907,0.12876,1.976\\
5,1536,15.237,0.19338,9.707,13.850,0.14530,7.064,22.107,0.08234,2.242,20.524,0.13851,2.242\\
6,2048,12.314,0.23932,12.520,10.201,0.24601,10.546,20.868,0.10066,2.920,18.942,0.17390,2.920\\
}\resolutioninteriorverse

\pgfplotsset{
  resolution axis/.style={
    scale only axis,
    width=289.92pt,
    height=256.32pt,
    anchor=north west,
    xmin=-0.364,
    xmax=6.361,
    axis lines=left,
    axis line style={draw=resolutionink!65,line width=.85pt,-},
    xtick={0,3,6},
    xticklabels={512,1024,2048},
    tick align=outside,
    tick style={draw=resolutionink!65,line width=.85pt},
    major tick length=3.2pt,
    xmajorgrids=false,
    ymajorgrids=true,
    major grid style={draw=resolutionink!12,line width=.65pt},
    tick label style={
      text=resolutionink,
      font=\sffamily\fontsize{22}{25}\selectfont
    },
    label style={
      text=resolutionink,
      font=\sffamily\fontsize{24}{28}\selectfont
    },
    ylabel style={xshift=-4pt},
    xlabel style={yshift=-6pt},
    yticklabel style={
      /pgf/number format/fixed,
      /pgf/number format/1000 sep={},
      /pgf/number format/assume math mode=true
    },
    clip=false
  }
}

\def\resolutionseries#1#2#3#4{%
  \addplot[
    color=#1,
    line width=2.5pt,
    line cap=round,
    line join=round,
    mark=*,
    scatter,
    visualization depends on={3.713551*sqrt(\thisrow{#2_memory}) \as \resolutionradius},
    scatter/use mapped color={draw=white,fill=#1},
    scatter/@pre marker code/.append style={/tikz/mark size=\resolutionradius pt},
    mark options={line width=1.2pt,fill opacity=.90,draw opacity=1}
  ] table[x=index,y=#3,col sep=comma] {#4};%
}

\resizebox{\linewidth}{!}{%
\begin{tikzpicture}[
  x=.24pt,
  y=-.24pt,
  font=\sffamily,
  every node/.style={text=resolutionink,inner sep=0pt}
]

\foreach \x/\y/\col/\txt in {
  205/106/resolutionblue/MVInverse,
  1087/106/resolutionorange/{RGB$\leftrightarrow$X},
  747/250/resolutionours/Ours,
  1350/250/resolutionpurple/{Ours w/o multi-res}
}{
  \draw[\col,line width=2.5pt,line cap=round]
    (\x,\y) -- ++(195,0);
  \draw[fill=\col,draw=white,line width=1.2pt]
    (\x+97.5,\y) circle[radius=4.5pt];
  \node[
    anchor=west,
    font=\sffamily\fontsize{24}{28}\selectfont
  ] at (\x+237,\y) {\txt};
}

\node[
  anchor=west,
  font=\sffamily\fontsize{22}{26}\selectfont
] at (1820,106) {Memory:};

\foreach \x/\r/\txt in {2290/5.252/2, 2450/9.096/6, 2640/12.864/{12\,GB}}{
  \fill[resolutionink!45] (\x,106) circle[radius=\r pt];
  \node[
    anchor=west,
    font=\sffamily\fontsize{22}{26}\selectfont
  ] at ($(\x,106)+(\r pt,0)+(14,0)$) {\txt};
}

\node[
  font=\sffamily\bfseries\fontsize{26}{30}\selectfont
] at (1730,481) {Hypersim};

\node[
  font=\sffamily\bfseries\fontsize{26}{30}\selectfont
] at (1730,1861) {InteriorVerse};

\coordinate (resolutionTL) at (296,559);
\coordinate (resolutionTR) at (1941,559);
\coordinate (resolutionBL) at (296,1939);
\coordinate (resolutionBR) at (1941,1939);

\begin{axis}[
  resolution axis,
  at={(resolutionTL)},
  ymin=10.847,
  ymax=17.375,
  ytick={11,13,15,17},
  xticklabels=\empty,
  ylabel={Albedo PSNR (dB) $\uparrow$}
]
  \resolutionseries
    {resolutionblue}{mvinverse}
    {mvinverse_psnr}{\resolutionhypersim}
  \resolutionseries
    {resolutionpurple}{single}
    {single_psnr}{\resolutionhypersim}
  \resolutionseries
    {resolutionours}{ours}
    {ours_psnr}{\resolutionhypersim}
\end{axis}

\begin{axis}[
  resolution axis,
  at={(resolutionTR)},
  ymin=.17032,
  ymax=.41835,
  ytick={.20,.25,.30,.35,.40},
  yticklabel style={
    /pgf/number format/fixed zerofill,
    /pgf/number format/precision=2
  },
  xticklabels=\empty,
  ylabel={Albedo LPIPS $\downarrow$}
]
  \resolutionseries
    {resolutionblue}{mvinverse}
    {mvinverse_lpips}{\resolutionhypersim}
  \resolutionseries
    {resolutionpurple}{single}
    {single_lpips}{\resolutionhypersim}
  \resolutionseries
    {resolutionours}{ours}
    {ours_lpips}{\resolutionhypersim}
\end{axis}

\begin{axis}[
  resolution axis,
  at={(resolutionBL)},
  ymin=8.0061,
  ymax=24.655,
  ytick={10,15,20},
  ylabel={Albedo PSNR (dB) $\uparrow$},
  xlabel={Input resolution (px)}
]
  \resolutionseries
    {resolutionblue}{mvinverse}
    {mvinverse_psnr}{\resolutioninteriorverse}
  \resolutionseries
    {resolutionorange}{rgbx}
    {rgbx_psnr}{\resolutioninteriorverse}
  \resolutionseries
    {resolutionpurple}{single}
    {single_psnr}{\resolutioninteriorverse}
  \resolutionseries
    {resolutionours}{ours}
    {ours_psnr}{\resolutioninteriorverse}
\end{axis}

\begin{axis}[
  resolution axis,
  at={(resolutionBR)},
  ymin=.04343,
  ymax=.27733,
  ytick={.05,.10,.15,.20,.25},
  yticklabel style={
    /pgf/number format/fixed zerofill,
    /pgf/number format/precision=2
  },
  ylabel={Albedo LPIPS $\downarrow$},
  xlabel={Input resolution (px)}
]
  \resolutionseries
    {resolutionblue}{mvinverse}
    {mvinverse_lpips}{\resolutioninteriorverse}
  \resolutionseries
    {resolutionorange}{rgbx}
    {rgbx_lpips}{\resolutioninteriorverse}
  \resolutionseries
    {resolutionpurple}{single}
    {single_lpips}{\resolutioninteriorverse}
  \resolutionseries
    {resolutionours}{ours}
    {ours_lpips}{\resolutioninteriorverse}
\end{axis}

\pgfresetboundingbox
\path[use as bounding box]
  (0,0) rectangle (3198,3229);

\end{tikzpicture}%
}

\endgroup
\caption{\textbf{Resolution robustness and memory scaling.}
Albedo PSNR~($\uparrow$) and LPIPS~($\downarrow$) on Hypersim
and InteriorVerse for input resolutions from $512$ to $2048$ pixels.
Marker area encodes peak inference GPU memory.
We compare our models trained with and without multi-resolution
supervision against MVInverse and RGB$\leftrightarrow$X (InteriorVerse only).
Our model is more stable across resolutions;
multi-resolution supervision further improves high-resolution
accuracy while keeping peak memory below $3$\,GB.}
\label{fig:resolution}
\end{figure}
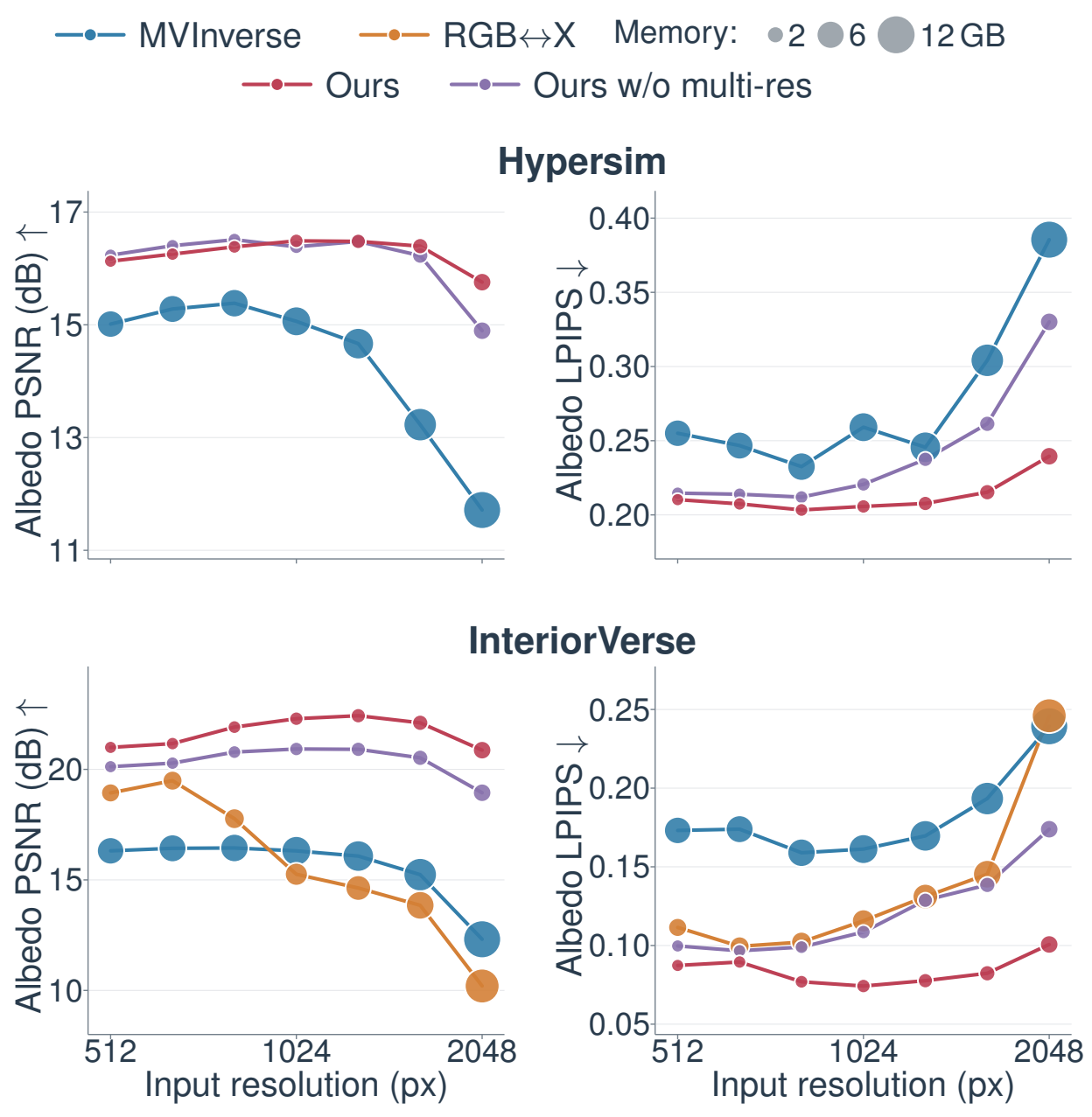

\myparagraph{Evaluation protocol and baselines.}
We evaluate on the official test splits of Hypersim~\cite{Roberts:2021:Hypersim} and InteriorVerse~\cite{Zhu:2022:InteriorVerse}, containing $7{,}690$ and $2{,}672$ images, respectively; for InteriorVerse, we use the $120^\circ$ FOV release from a complete copy of the dataset obtained before its 2026 data loss. Hypersim provides ground truth for albedo, irradiance, and surface normals, whereas InteriorVerse provides albedo, surface normals, roughness, and metallicity. All predictions are evaluated at the native resolution of each dataset ($1024{\times}768$ for Hypersim, $640{\times}480$ for InteriorVerse); outputs of baselines with a fixed inference resolution are resized accordingly. We report PSNR and RMSE in linear RGB and LPIPS after gamma encoding for albedo and irradiance; PSNR, RMSE, and MAE for roughness and metallicity; and, for unit-normalized normals, the mean angular error and the percentage of pixels below $11.25^\circ$ and $30^\circ$. Metrics use valid pixels only; for LPIPS, invalid regions are filled with ground truth. We compare against ColorfulDiffuse~\cite{Careaga:2024:ColorfulIID}, IntrinsicImageDiffusion~\cite{Kocsis:2024:IntrinsicImageDiffusion}, DiffusionRenderer~\cite{Liang:2025:DiffusionRenderer}, RGB$\leftrightarrow$X~\cite{Zeng:2024:rgbx}, and MVInverse~\cite{wu:2025:MVInverse}. Each baseline is evaluated only on the modalities it predicts.
Further details are provided in the supplementary material.

\begin{figure}[t]
  \centering
  \includegraphics[width=\linewidth]{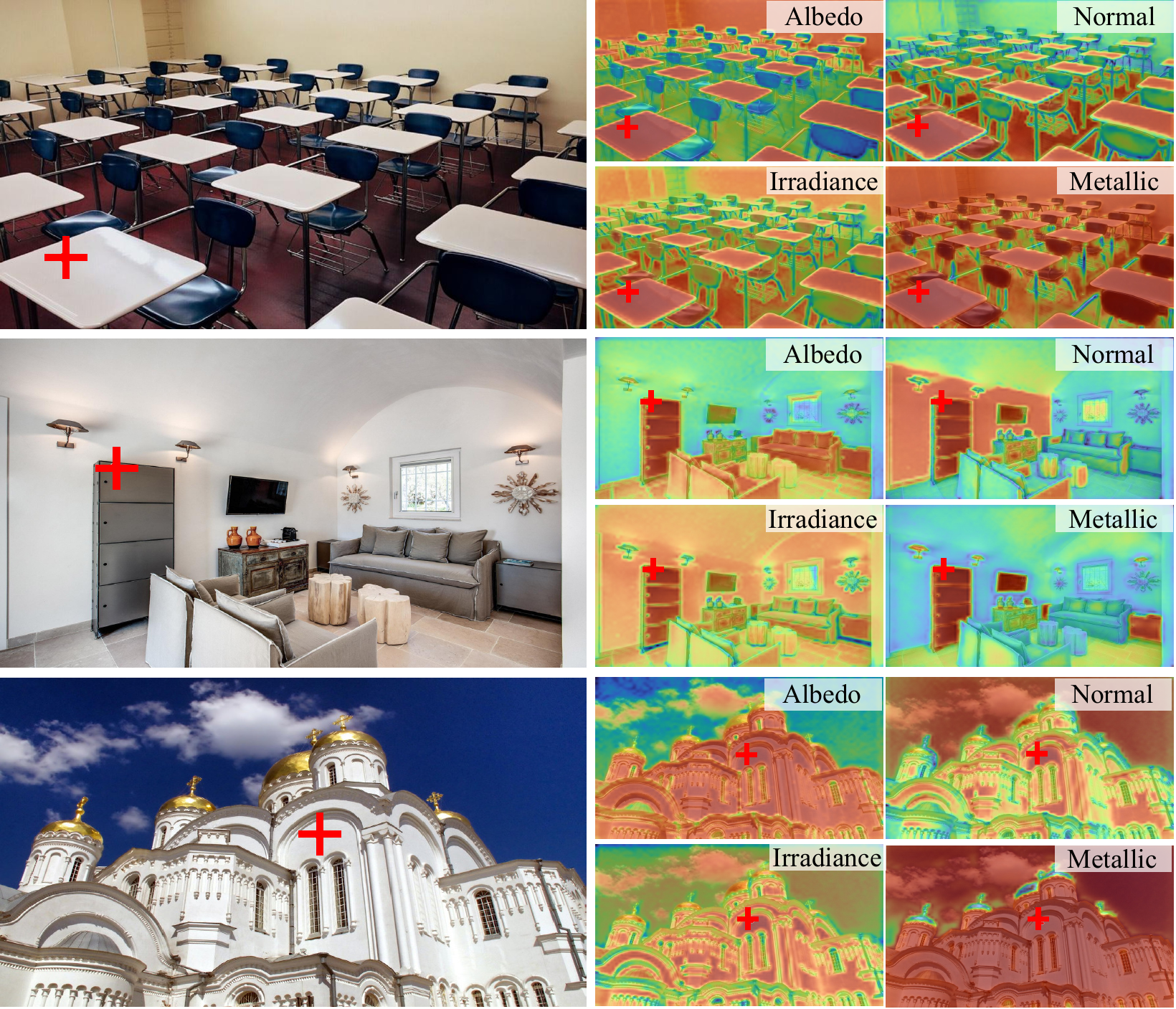}
  \caption{\textbf{Feature-space selection.}
  Given an image and a query patch \textcolor{red}{+}, we compute cosine similarity between the query and patch features in each modality-projected feature space (red: high, blue: low). The resulting maps show that albedo, normal, irradiance, and metallicity projections induce different selection axes, grouping regions according to the corresponding physical property.}
  \label{fig:selection}
  \vspace{-0.8em}
\end{figure}

\begin{figure*}[t]
\centering
\begingroup

\definecolor{figureInk}{HTML}{293B4C}
\definecolor{figureMuted}{HTML}{586673}

\tikzset{
  comparisonPanel/.style={
    anchor=north west,
    inner sep=0pt,
    outer sep=0pt
  },
  comparisonHeading/.style={
    align=center,
    text=figureInk,
    font=\rmfamily\fontsize{7.2}{8.6}\selectfont,
    inner sep=0pt,
    outer sep=0pt
  },
  resolutionBand/.style={
    anchor=center,
    rotate=90,
    text=black,
    font=\normalfont\rmfamily\fontsize{5.6}{6.7}\selectfont,
    inner sep=0pt,
    outer sep=0pt
  }
}

\def\cw{1.53cm}

\resizebox{\textwidth}{!}{%
\begin{tikzpicture}[
  x=0.017cm,
  y=-0.017cm,
  every node/.style={inner sep=0pt,outer sep=0pt}
]

\path[use as bounding box]
  (230,132) rectangle (1152,525);

\node[comparisonPanel] at (402,162) {\includegraphics[width=\cw,height=\cw]{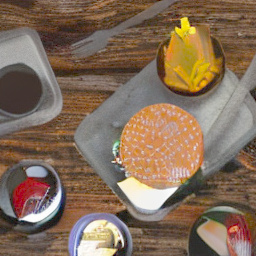}};
\node[comparisonPanel] at (492,162) {\includegraphics[width=\cw,height=\cw]{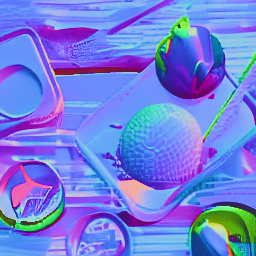}};

\node[comparisonPanel] at (402,253) {\includegraphics[width=\cw,height=\cw]{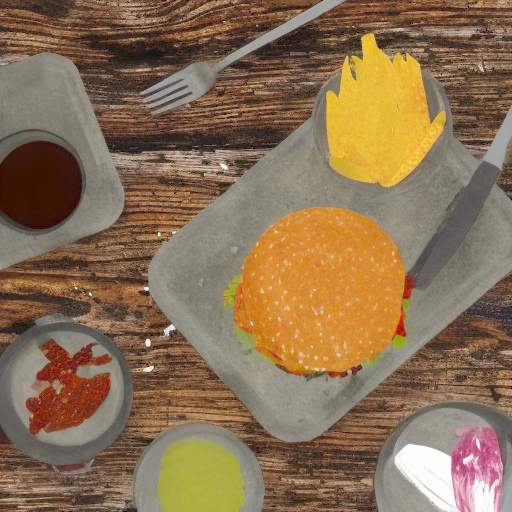}};
\node[comparisonPanel] at (492,253) {\includegraphics[width=\cw,height=\cw]{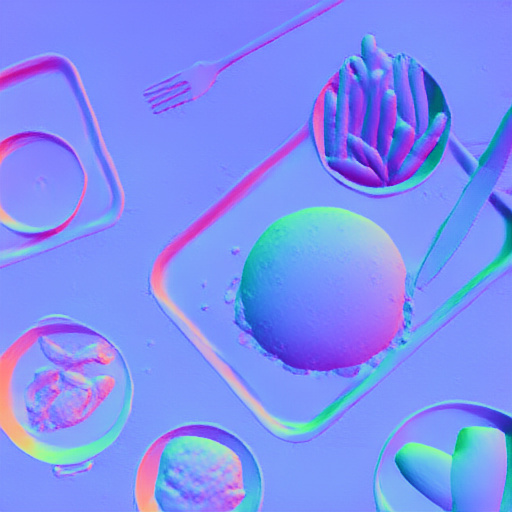}};

\node[comparisonPanel] at (402,344) {\includegraphics[width=\cw,height=\cw]{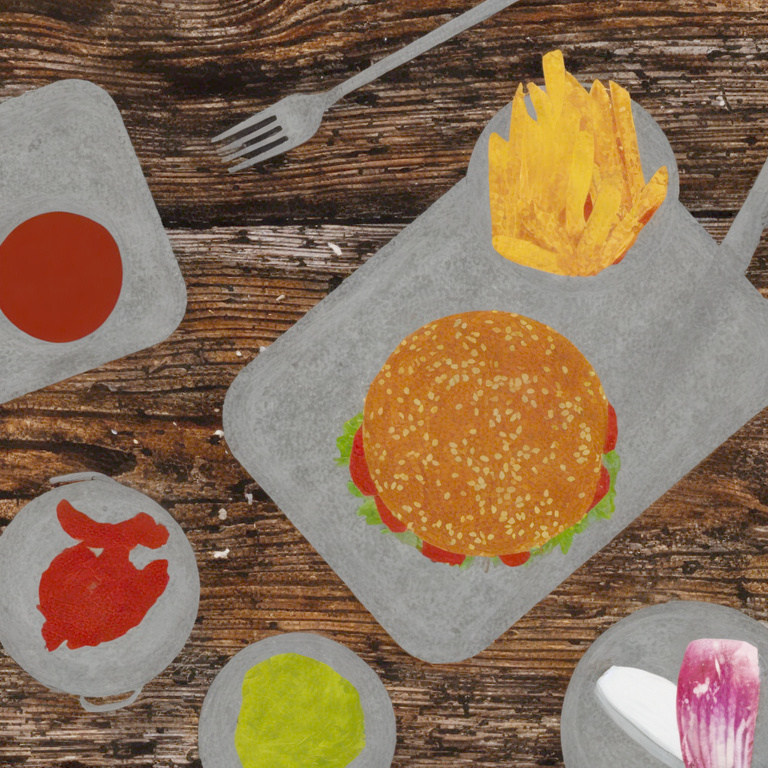}};
\node[comparisonPanel] at (492,344) {\includegraphics[width=\cw,height=\cw]{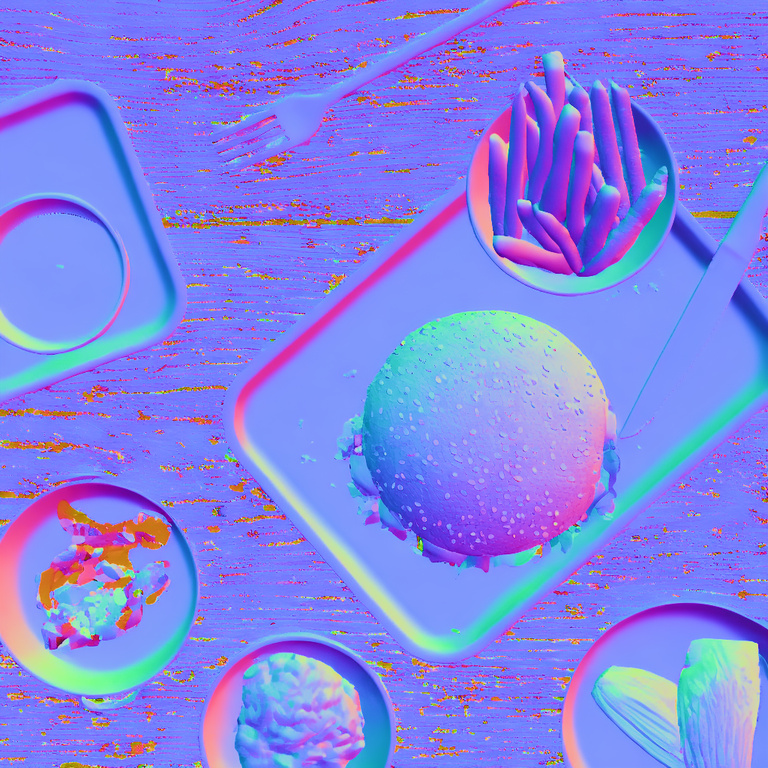}};

\node[comparisonPanel] at (402,435) {\includegraphics[width=\cw,height=\cw]{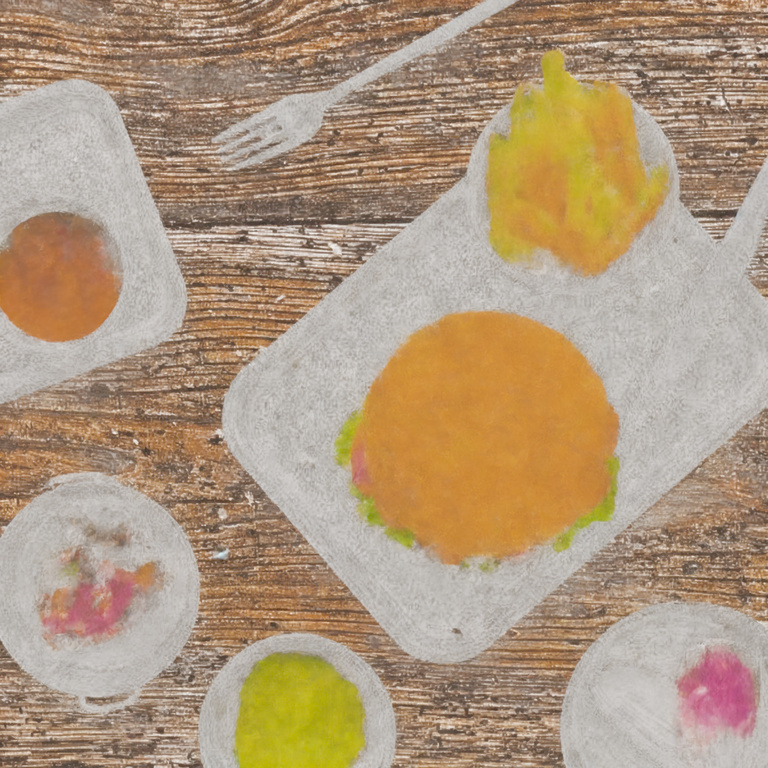}};
\node[comparisonPanel] at (492,435) {\includegraphics[width=\cw,height=\cw]{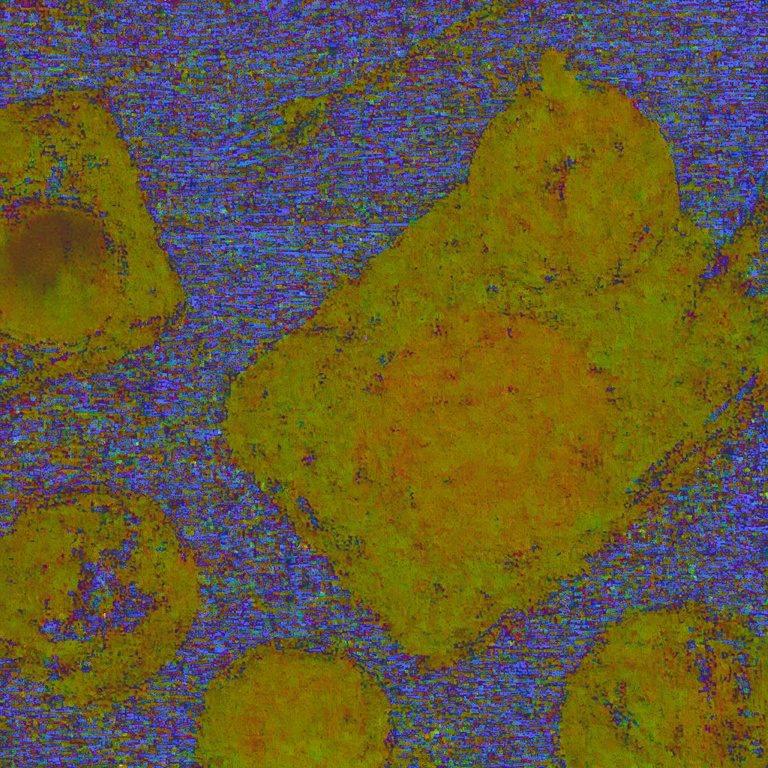}};

\node[comparisonPanel] at (592,162) {\includegraphics[width=\cw,height=\cw]{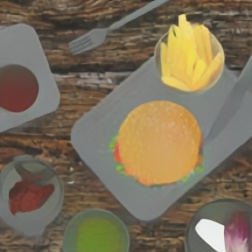}};
\node[comparisonPanel] at (682,162) {\includegraphics[width=\cw,height=\cw]{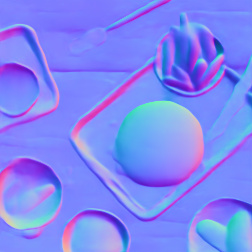}};

\node[comparisonPanel] at (592,253) {\includegraphics[width=\cw,height=\cw]{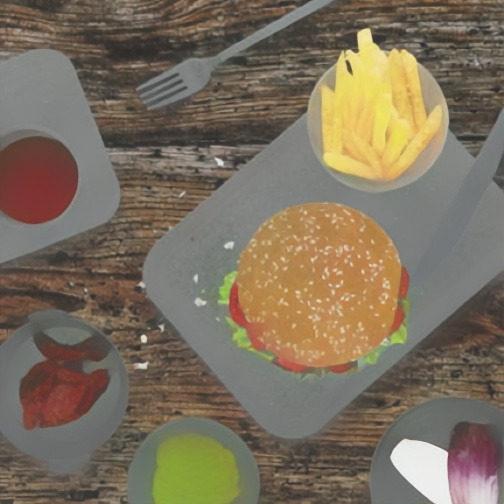}};
\node[comparisonPanel] at (682,253) {\includegraphics[width=\cw,height=\cw]{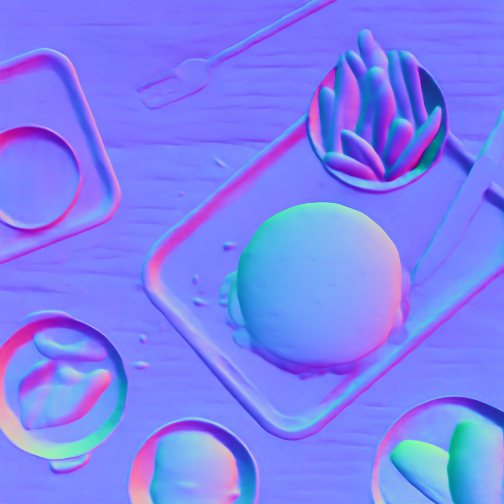}};

\node[comparisonPanel] at (592,344) {\includegraphics[width=\cw,height=\cw]{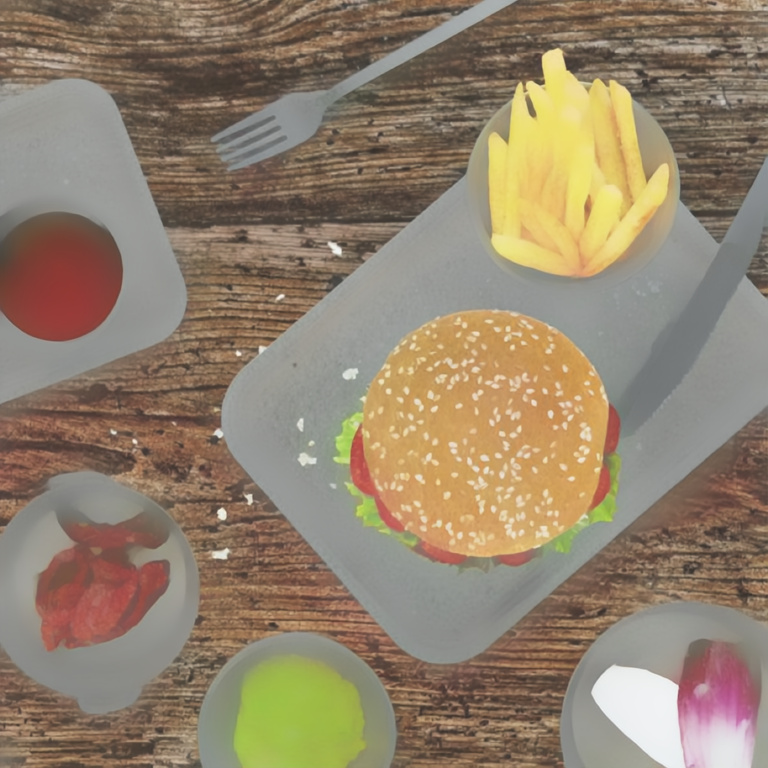}};
\node[comparisonPanel] at (682,344) {\includegraphics[width=\cw,height=\cw]{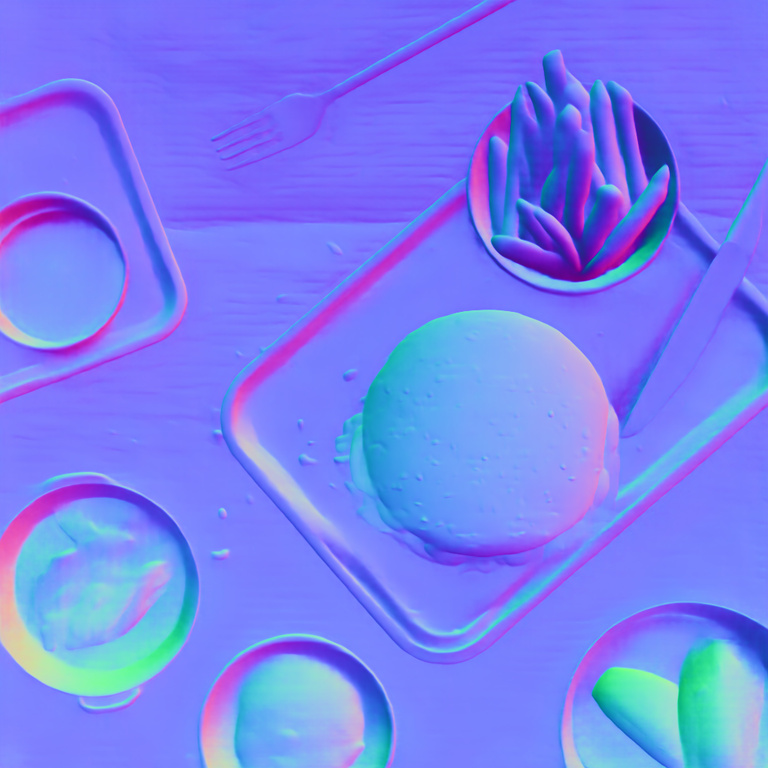}};

\node[comparisonPanel] at (592,435) {\includegraphics[width=\cw,height=\cw]{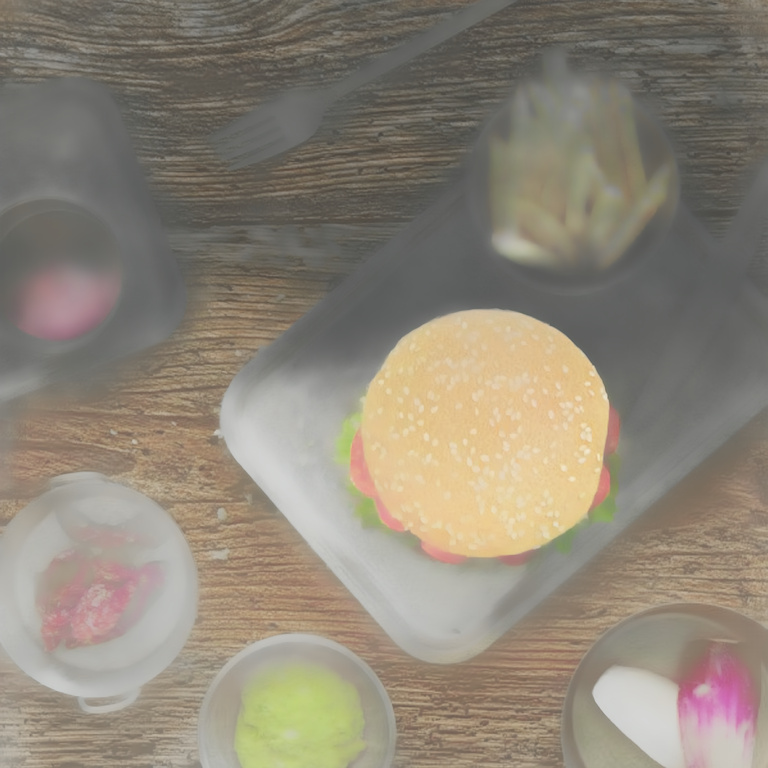}};
\node[comparisonPanel] at (682,435) {\includegraphics[width=\cw,height=\cw]{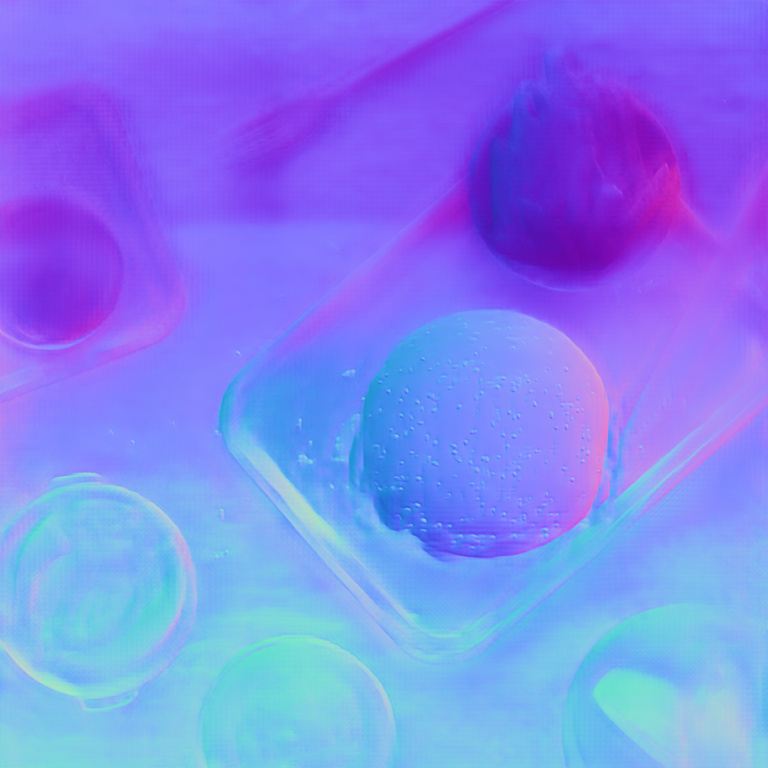}};

\node[comparisonPanel] at (230,281) {%
  \includegraphics[width=2.21cm,height=2.21cm]
    {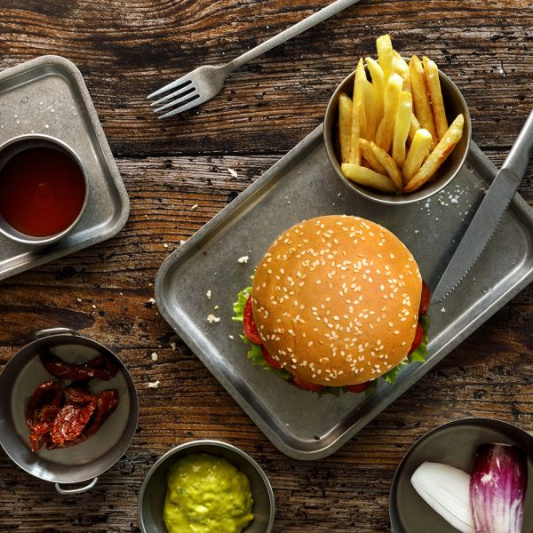}
};

\node[
  align=center,
  text=figureMuted,
  font=\rmfamily\fontsize{7.2}{8.6}\selectfont
] at (295,428) {Input RGB};

\node[comparisonPanel] at (782,162) {\includegraphics[width=\cw,height=\cw]{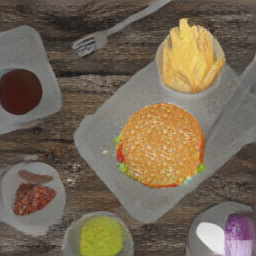}};
\node[comparisonPanel] at (872,162) {\includegraphics[width=\cw,height=\cw]{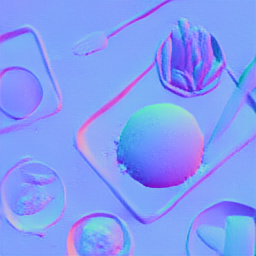}};

\node[comparisonPanel] at (782,253) {\includegraphics[width=\cw,height=\cw]{images/images_pagina7/panel_25.jpg}};
\node[comparisonPanel] at (872,253) {\includegraphics[width=\cw,height=\cw]{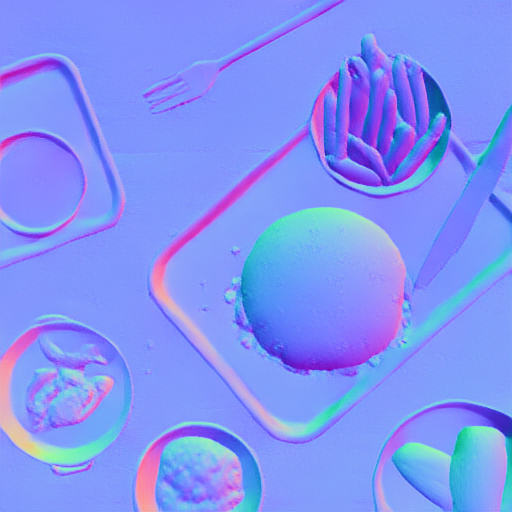}};

\node[comparisonPanel] at (782,344) {\includegraphics[width=\cw,height=\cw]{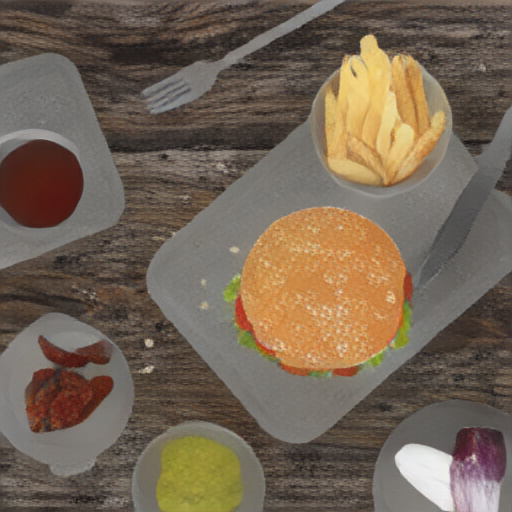}};
\node[comparisonPanel] at (872,344) {\includegraphics[width=\cw,height=\cw]{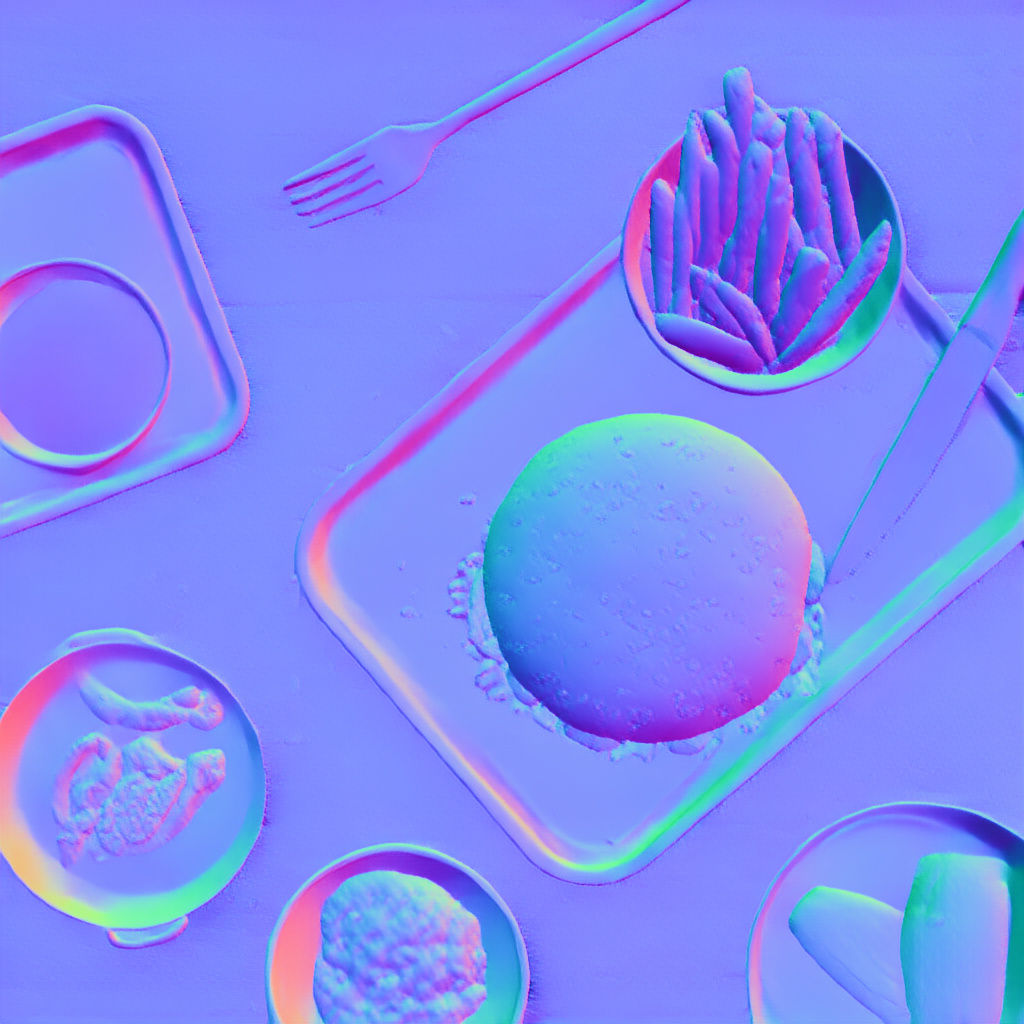}};

\node[comparisonPanel] at (782,435) {\includegraphics[width=\cw,height=\cw]{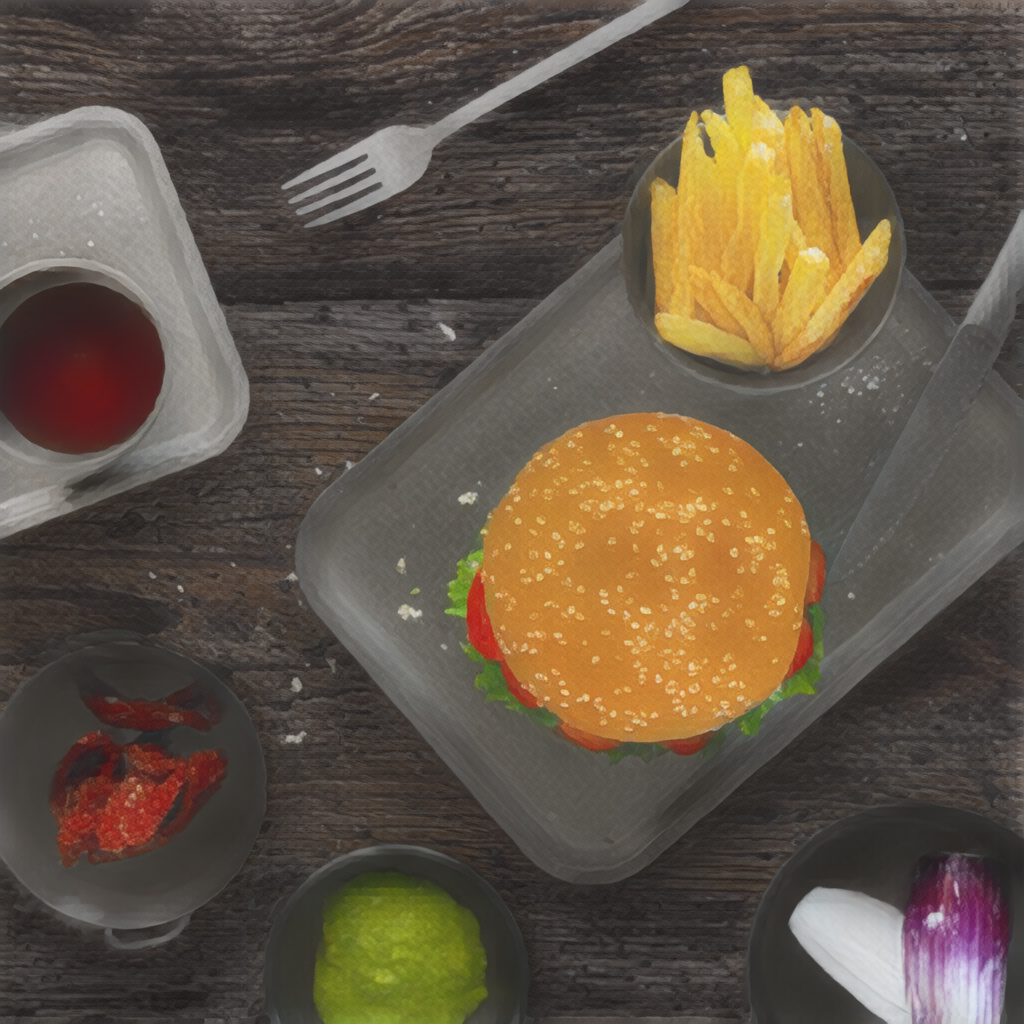}};
\node[comparisonPanel] at (872,435) {\includegraphics[width=\cw,height=\cw]{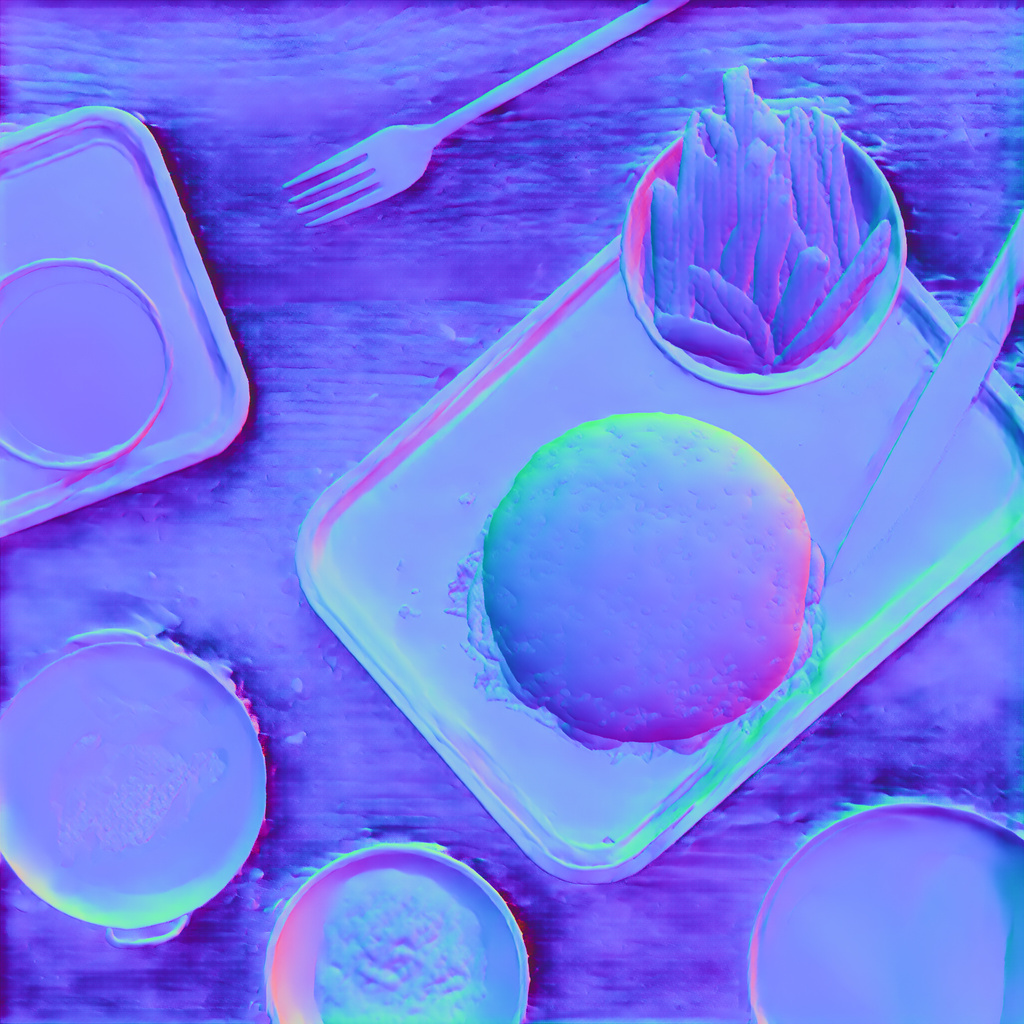}};

\node[comparisonPanel] at (972,162) {\includegraphics[width=\cw,height=\cw]{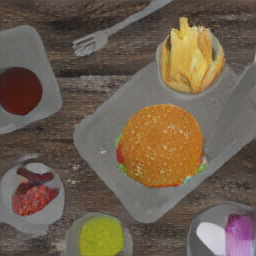}};
\node[comparisonPanel] at (1062,162) {\includegraphics[width=\cw,height=\cw]{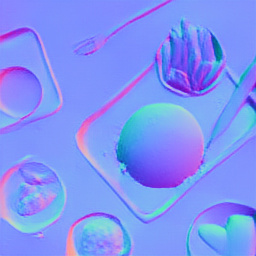}};

\node[comparisonPanel] at (972,253) {\includegraphics[width=\cw,height=\cw]{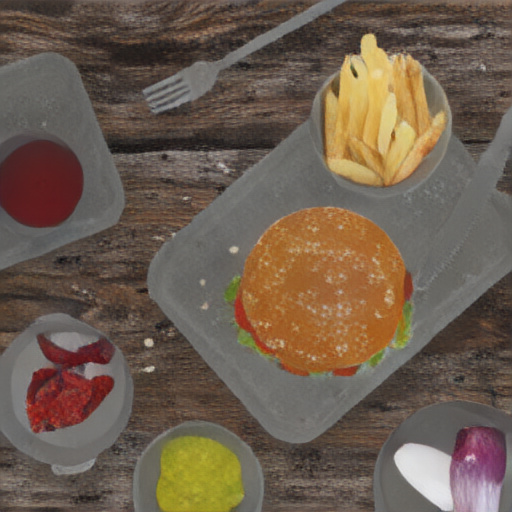}};
\node[comparisonPanel] at (1062,253) {\includegraphics[width=\cw,height=\cw]{images/images_pagina7/panel_04.jpg}};

\node[comparisonPanel] at (972,344) {\includegraphics[width=\cw,height=\cw]{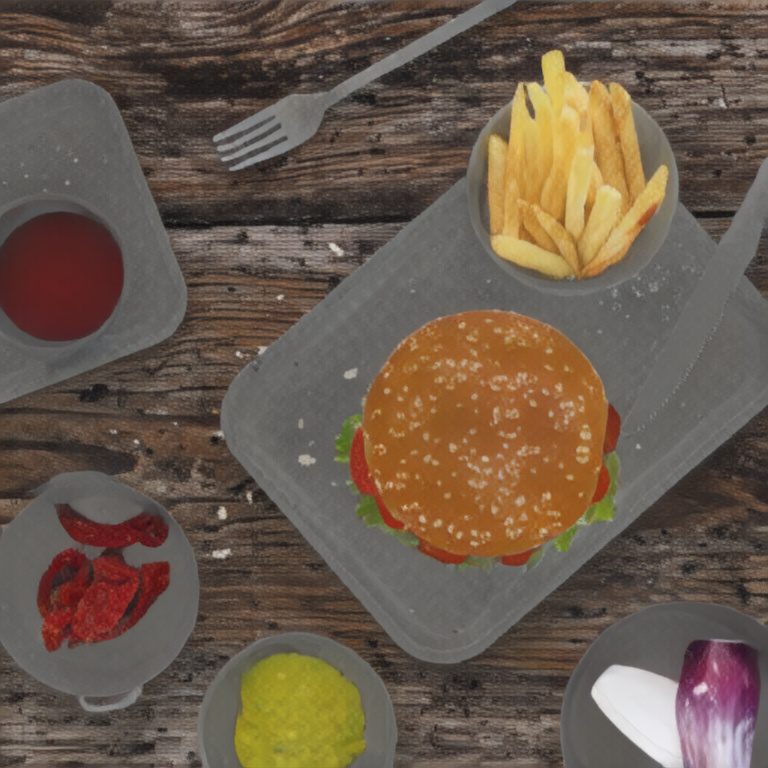}};
\node[comparisonPanel] at (1062,344) {\includegraphics[width=\cw,height=\cw]{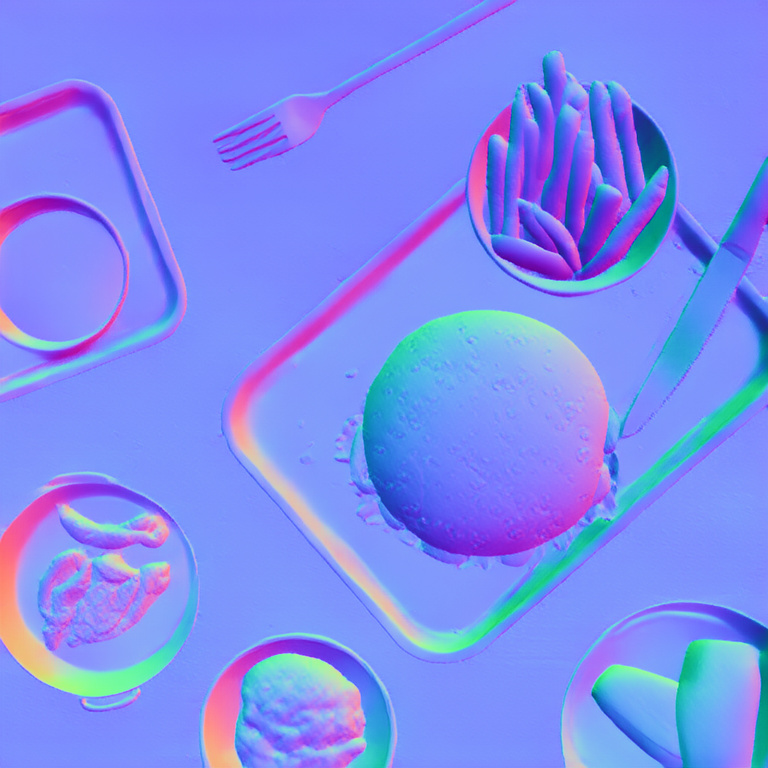}};

\node[comparisonPanel] at (972,435) {\includegraphics[width=\cw,height=\cw]{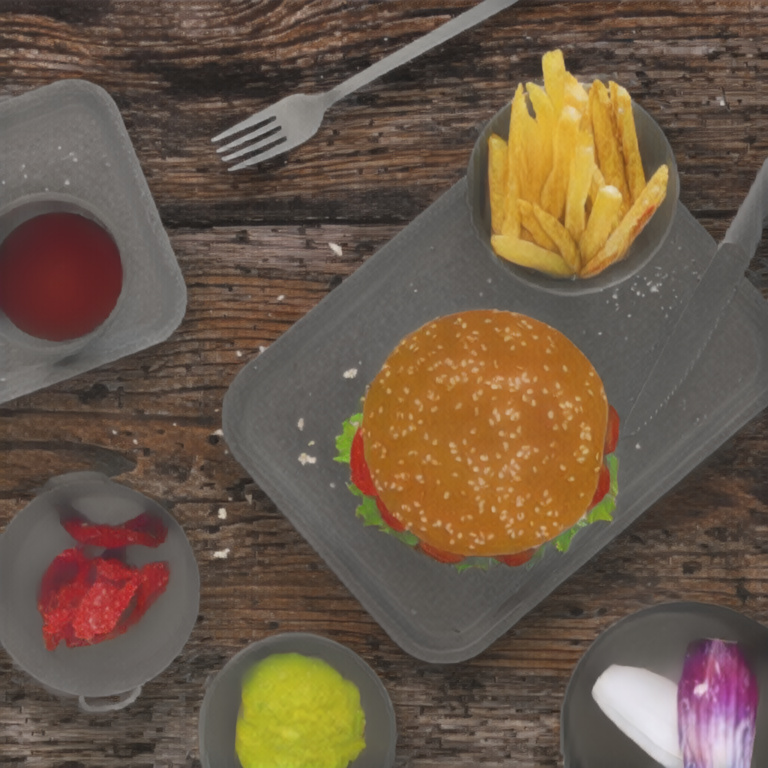}};
\node[comparisonPanel] at (1062,435) {\includegraphics[width=\cw,height=\cw]{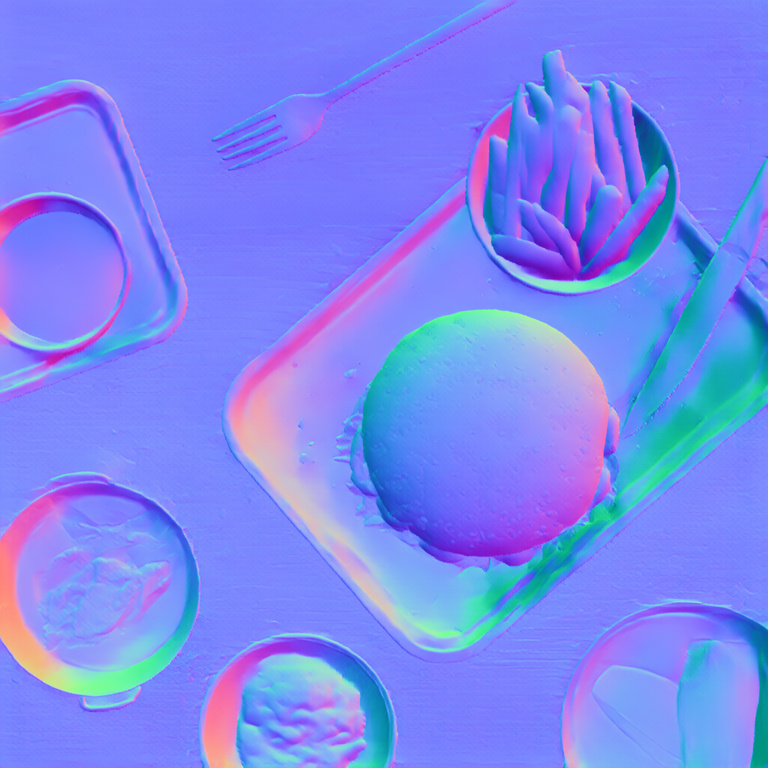}};

\node[comparisonHeading]
  at (492,146) {RGB$\rightarrow$X};

\node[comparisonHeading]
  at (682,146) {MVInverse};

\node[comparisonHeading]
  at (872,146) {\methodName{} w/o multi-res. training};

\node[
  comparisonHeading,
  font=\rmfamily\bfseries\fontsize{7.2}{8.6}\selectfont
] at (1062,146) {\methodName{} (ours)};

\node[resolutionBand] at (389,207) {$256{\times}256$};
\node[resolutionBand] at (389,298) {$512{\times}512$};
\node[resolutionBand] at (389,389) {$1024{\times}1024$};
\node[resolutionBand] at (389,480) {$2048{\times}2048$};

\end{tikzpicture}%
}

\endgroup
\vspace{-0.1em}

\caption{\textbf{Qualitative robustness to input resolution.}
We compare RGB$\leftrightarrow$X, MVInverse, and \methodName{} with and without multi-resolution training at input resolutions from $256{\times}256$ to $2048{\times}2048$. At each resolution, we show the predicted albedo and surface normal maps.
RGB$\leftrightarrow$X exhibits artifacts at both extremes, whereas MVInverse degrades at high resolution.
The single-resolution variant of \methodName{} produces more consistent predictions than either baseline, suggesting that our adaptation to intrinsic decomposition preserves the resolution robustness of the pretrained DINOv3 representation. Multi-resolution training further improves prediction consistency.}
\label{fig:resolution_robustness}
\vspace{-0.3em}
\end{figure*}

\subsection{Main results}

\myparagraph{Intrinsic decomposition.}
\Cref{tab:intrinsic_quantitative} compares \methodName{} with diffusion-based and feed-forward methods. On InteriorVerse, our model achieves the best roughness and metallicity on all metrics ($19.67$ and $20.82$ dB PSNR), the lowest mean normal error ($8.55^\circ$), and the best albedo LPIPS; its albedo PSNR ($20.92$ dB) is second only to DiffusionRenderer~\cite{Liang:2025:DiffusionRenderer} ($21.19$ dB), which in turn performs poorly on materials. On Hypersim, \methodName{} achieves the best irradiance reconstruction, exceeding MVInverse~\cite{wu:2025:MVInverse} by $2.43$ dB in PSNR. It also achieves the lowest albedo RMSE and LPIPS, while MVInverse retains a slight advantage in albedo PSNR ($0.26$ dB) and normals.
Notably, \methodName{} achieves this with the lowest parameter count and highest inference throughput among the evaluated methods. With $165$M parameters, it uses approximately one-sixth as many parameters as MVInverse~\cite{wu:2025:MVInverse}, while outperforming it on every metric, except normal estimation. It is also $2.7{\times}$ smaller than ColorfulDiffuse~\cite{Careaga:2024:ColorfulIID} and at least $7.8{\times}$ smaller than the evaluated diffusion backbones. This compact architecture translates into higher inference throughput: \methodName{} processes $14.97$ FPS at $640{\times}480$ and $8.94$ FPS at $1024{\times}768$, compared with $8.96$ and $4.45$ FPS for MVInverse, respectively. On InteriorVerse, it exceeds the throughput of the fastest diffusion baseline, RGB$\leftrightarrow$X~\cite{Zeng:2024:rgbx}, by over $100{\times}$. These results show that a compact feed-forward model can match or surpass substantially larger models in appearance, material, and illumination estimation.
\Cref{fig:teaser} shows qualitative results on real images; more results in the supplement.

\myparagraph{Scalability across resolutions.}
\Cref{fig:resolution} evaluates robustness to input resolution on fixed $500$-image subsets of both test sets, rescaling the long side from $512$ to $2048$ pixels; since most sizes exceed the native resolution, this measures robustness to rescaling rather than recovery of additional detail. Ours is substantially more stable than prior methods: on InteriorVerse, albedo PSNR drops only from $20.99$ to $20.86$ dB, versus $16.31$ to $12.31$ dB for MVInverse and $18.93$ to $10.19$ dB for RGB$\leftrightarrow$X; on Hypersim, it drops by $0.37$ dB versus $3.30$ dB for MVInverse. Peak memory grows only from $1.46$ to $2.92$ GB, compared with $6.50$--$12.53$ GB for MVInverse and $3.10$--$10.53$ GB for RGB$\leftrightarrow$X.
Even our single-resolution variant degrades less than the baselines ($1.18$ and $1.34$ dB), suggesting that the frozen DINOv3 representation already provides much of this robustness; multi-resolution supervision adds $1.93$ and $0.86$ dB at $2048$ pixels on InteriorVerse and Hypersim. \Cref{fig:resolution_robustness} shows the same trend qualitatively.

\myparagraph{Modality-specific feature selection.}
To examine whether the projections yield modality-specific representations, we perform similarity-based selection in each projected feature space: given a query patch, we compute its cosine similarity to all other patches after each projection head. \Cref{fig:selection} shows that similarity follows the physical quantity of each modality. In the top example, a query on a white non-metallic desk retrieves similarly colored surfaces in the albedo space, similarly oriented planes in the normal space, similarly lit regions in the irradiance space, and non-metallic regions in the metallicity space, suppressing metallic chair armrests and table legs. Feature alignment thus yields not only decodable features, but feature spaces organized by the physical quantity of each modality.

\myparagraph{Choice of the pretrained representation.}
Intrinsic decomposition requires a representation that is both semantically rich and spatially precise. Our hypothesis is that DINOv3~\cite{Simeoni:2025:Dinov3} provides this combination: it retains the strong semantic abstraction of earlier self-supervised DINO models~\cite{Caron:2021:DINO,Oquab:2023:Dinov2}, while producing high-quality, spatially smooth dense features that are particularly suitable for dense prediction~\cite{Cuttano:2026:INSID3}. To test this hypothesis, we replace DINOv3 with alternative visual backbones while keeping the rest of the architecture and training protocol unchanged. As shown in \cref{tab:backbone_ablation_hypersim}, DINOv3 achieves the best performance across all evaluated modalities, consistently improving over DINOv2 \cite{Oquab:2023:Dinov2}, PE-Core \cite{Bolya:2025:PerceptionEncoder}, PE-Spatial \cite{Bolya:2025:PerceptionEncoder}, and SigLIP \cite{Tschannen:2025:SigLIP2}. 

\myparagraph{Training signals.}
\Cref{tab:signal_ablation} ablates the role of the reconstruction teacher and the output loss in intrinsic prediction. We compare three variants. \emph{No teacher} predicts each intrinsic map directly in pixel space with a separate trainable decoder, using output supervision only. \emph{No pixel loss} relies solely on latent prediction, training the student to match the teacher features, without supervising the decoded maps. \methodName{} combines feature alignment with output supervision through the frozen decoder. 
Predicting directly in pixel space (\emph{No teacher}) achieves good accuracy on albedo and normals, but fails on the most ambiguous properties: compared to \methodName{}, metallicity and roughness PSNR drop by $9.1$ and $11.6$ dB on InteriorVerse, and irradiance PSNR by $3.1$ dB on Hypersim. Feature alignment alone (\emph{No pixel loss}) largely recovers material prediction ($+7.0$ and $+6.3$ dB over direct regression) and improves irradiance, suggesting that the teacher space makes these properties easier to infer, but degrades albedo and normals, whose fine details are no longer supervised directly. Adding the pixel loss improves every metric over feature alignment alone, yielding the best normals, metallicity, and roughness on InteriorVerse and the best irradiance on Hypersim. For albedo, the property closest to the RGB input and thus easier to regress directly, \methodName{} achieves higher PSNR than direct pixel prediction on InteriorVerse but lower on Hypersim. 
In the supplement, we further analyze the contributions of teacher reconstruction and latent prediction to the overall error.

\begin{table}[t]
\centering
\caption{\textbf{Choice of pretrained representation.}
We compare DINOv3 with leading Vision Foundation Models spanning self-supervised and image--text pretraining. With each VFM, we train a reconstruction autoencoder and then an RGB student to predict the teacher features. DINOv3 consistently achieves the best performance across all modalities on Hypersim. Best result in \textbf{bold}.}
\label{tab:backbone_ablation_hypersim}
\vspace{-0.3em}
\footnotesize
\setlength{\tabcolsep}{3pt}
\resizebox{\columnwidth}{!}{%
\begin{tabular}{@{}lcc|cc|cc@{}}
\toprule
&
\multicolumn{2}{c|}{\textbf{Albedo}} &
\multicolumn{2}{c|}{\textbf{Normal}} &
\multicolumn{2}{c}{\textbf{Irradiance}} \\
\cmidrule(lr){2-3}
\cmidrule(lr){4-5}
\cmidrule(lr){6-7}
\textbf{Backbone} &
PSNR $\uparrow$ &
LPIPS $\downarrow$ &
Mean $\downarrow$ &
$11.25^\circ$ $\uparrow$ &
PSNR $\uparrow$ &
LPIPS $\downarrow$ \\
\midrule

PE-Core \cite{Bolya:2025:PerceptionEncoder}
& 14.68 & 0.31
& 39.10 & 20.75
& 16.02 & 0.30 \\

PE-Spatial \cite{Bolya:2025:PerceptionEncoder}
& 14.27 & 0.29
& 34.01 & 25.06
& 16.16 & 0.28 \\

SigLIP \cite{Tschannen:2025:SigLIP2}
& 15.01 & 0.28
& 27.96 & 36.08
& 16.56 & 0.28 \\

DINOv2 \cite{Oquab:2023:Dinov2}
& 15.61 & 0.26
& 19.73 & 56.19
& 17.77 & 0.26 \\

\textbf{DINOv3} \cite{Simeoni:2025:Dinov3}
& \textbf{16.32} & \textbf{0.21}
& \textbf{16.90} & \textbf{66.79}
& \textbf{17.86} & \textbf{0.23} \\

\bottomrule
\end{tabular}%
}
\vspace{-0.8em}
\vspace{-0.8em}
\end{table}
\begin{table}[t]
\centering

\caption{\textbf{Ablation of the reconstruction teacher and output loss.}
\emph{No teacher} regresses maps directly in pixel space, while \emph{No pixel loss} uses teacher feature alignment alone.
Combining feature alignment with supervision on the decoded maps (\methodName{}) gives the strongest overall performance, leading across all four modalities on InteriorVerse and on irradiance on Hypersim.
Direct regression achieves higher albedo PSNR and slightly lower mean normal error on Hypersim at the cost of a separate decoder per modality.}
\vspace{-0.3em}

\label{tab:signal_ablation}
\footnotesize
\setlength{\tabcolsep}{3pt}
\begin{tabularx}{\columnwidth}{@{}l*{4}{>{\centering\arraybackslash}X}|*{3}{>{\centering\arraybackslash}X}@{}}
\toprule
& \multicolumn{4}{c|}{\textbf{InteriorVerse}} & \multicolumn{3}{c}{\textbf{Hypersim}} \\
\cmidrule(lr){2-5}\cmidrule(lr){6-8}
\textbf{Variant} & Albedo & Normal & Metal. & Rough. & Albedo & Normal & Irr. \\
& {\tiny\mbox{PSNR $\uparrow$}} & {\tiny\mbox{Mean $\downarrow$}} & {\tiny\mbox{PSNR $\uparrow$}} & {\tiny\mbox{PSNR $\uparrow$}} & {\tiny\mbox{PSNR $\uparrow$}} & {\tiny\mbox{Mean $\downarrow$}} & {\tiny\mbox{PSNR $\uparrow$}} \\
\midrule
No teacher
& \underline{20.13} & \underline{8.70} & 11.73 & 8.10
& \textbf{18.33} & \textbf{16.22} & 14.76 \\
No pixel loss
& 19.34 & 10.30 & \underline{18.76} & \underline{14.41}
& 13.70 & 19.46 & \underline{16.79} \\
\midrule
\textbf{\methodName}
& \textbf{20.92} & \textbf{8.55} & \textbf{20.82} & \textbf{19.67}
& \underline{16.32} & \underline{16.90} & \textbf{17.86} \\
\bottomrule
\end{tabularx}
\vspace{-0.9em}
\end{table}

\section{Conclusion}

We presented \methodName{}, a compact model for single-image intrinsic decomposition that separates learning to reconstruct intrinsic maps from learning to infer them from RGB. The reconstruction teacher learns to recover intrinsic maps from DINOv3 features with a shared decoder, and the RGB student predicts the corresponding features for each modality. With $165$M parameters, \methodName{} estimates reflectance, illumination, geometry, and material properties in a single deterministic forward pass. It achieves the strongest material estimates among the evaluated methods and produces more consistent predictions as input resolution varies.
Future work could focus on extending the student to multiple views or video for cross-view and temporal consistency. Transparent surfaces such as windows might benefit from layered predictions separating the surface from the scene behind it via a transparency map, which the shared reconstruction space could host as new modalities. A rendering loss could encourage the maps to jointly explain the image.

\bibliographystyle{ieeenat_fullname}
\bibliography{bibtex/short,bibtex/references}

\appendix
\maketitlesupplementary

\noindent In this appendix, we provide additional analyses, implementation details, and experiments.

\begin{table*}[t]
\centering
\caption{\textbf{Data statistics.}
Number of training and test images, together with the available intrinsic-map annotations for each dataset.}
\label{tab:training_datasets}
\vspace{-0.5em}
\small
\setlength{\tabcolsep}{8pt}
\begin{tabular}{@{}l r r c c c c c@{}}
\toprule
\textbf{Dataset} &
\textbf{Train images} &
\textbf{Test images} &
\textbf{Albedo} &
\textbf{Irradiance} &
\textbf{Surface normals} &
\textbf{Roughness} &
\textbf{Metallicity} \\
\midrule

ABO~\cite{Collins:2022:ABO}
& $2{,}061{,}787$
& --
& \checkmark & -- & \checkmark & \checkmark & \checkmark \\

Evermotion~\cite{Zeng:2024:rgbx}
& $17{,}000$
& --
& \checkmark & -- & \checkmark & \checkmark & \checkmark \\

Hypersim~\cite{Roberts:2021:Hypersim}
& $59{,}542$
& $7{,}690$
& \checkmark & \checkmark & \checkmark & -- & -- \\

InteriorVerse~\cite{Zhu:2022:InteriorVerse}
& $50{,}097$
& $2{,}672$
& \checkmark & -- & \checkmark & \checkmark & \checkmark \\

MatrixCity~\cite{Li:2023:MatrixCity}
& $40{,}678$
& --
& \checkmark & -- & \checkmark & \checkmark & \checkmark \\

Objaverse~\cite{deitke:2022:objaverse}
& $88{,}700$
& --
& \checkmark & \checkmark & \checkmark & \checkmark & \checkmark \\

\midrule
\textbf{Total}
& $\mathbf{2{,}317{,}804}$
& $\mathbf{10{,}362}$
& & & & & \\
\bottomrule
\end{tabular}
\vspace{-0.8em}
\end{table*}

\section{Training Data Details}

Our training corpus contains $2.32$M rendered images from six datasets covering complementary scene domains, as summarized in \cref{tab:training_datasets}. ABO~\cite{Collins:2022:ABO} and Objaverse~\cite{deitke:2022:objaverse} provide diverse object-centric assets, including household products and long-tail 3D objects. For Objaverse, we use an $88{,}700$-image subset of the full dataset. Evermotion~\cite{Zeng:2024:rgbx}, Hypersim~\cite{Roberts:2021:Hypersim}, and InteriorVerse~\cite{Zhu:2022:InteriorVerse} contribute photorealistic indoor scenes with detailed appearance, geometry, illumination, and material annotations, while MatrixCity~\cite{Li:2023:MatrixCity} extends the training distribution to large-scale urban environments and outdoor viewpoints. Together, these datasets provide broad variation in object categories, scene layout, illumination, geometry, and material properties. We sample datasets uniformly when constructing each batch. For each sampled image, we apply supervision only to the intrinsic modalities available in its source dataset. For InteriorVerse, we train on the official training and validation scenes, which are disjoint from the test scenes.

\section{Baseline Evaluation Details}

\myparagraph{Metric standardization.}
We evaluate all baselines using their publicly released checkpoints on identical frozen test manifests using a single evaluator. Hypersim uses the 7,690-image official v1 scene-level test split at $1024{\times}768$. InteriorVerse uses the official test split (210 scenes, 2,672 images) at $120^\circ$ FOV and $640{\times}480$, from a complete copy of the dataset obtained before the 2026 data loss, with denoised RGB, geometric-mean exposure normalization to 0.18, and gamma-2.2 encoding. All methods receive the same prepared observation before model-specific preprocessing. Predictions are bilinearly resized to ground-truth resolution and evaluated under the same dataset validity mask. We compute appearance PSNR and RMSE in linear RGB, AlexNet LPIPS after gamma-2.2 encoding, angular normal metrics after unit normalization, and scalar roughness and metallicity errors on the unit interval. Albedo is evaluated without any scale alignment. Metrics are computed per image and averaged over the test set. We use seed 0 and each method's released inference policy: one 50-step sample per RGB-X map, the image-space mean of ten 50-step Intrinsic Image Diffusion samples, one 20-step stochastic DiffusionRenderer sample per map, and deterministic feed-forward inference for MVInverse, ColorfulDiffuse, and \methodName.

\myparagraph{Efficiency protocol.}
We report end-to-end throughput for one input image under each method's complete inference protocol. Timing includes preprocessing, all map-specific passes and stochastic samples, postprocessing, data transfers, and output resizing, while excluding model and dataset loading. Measurements use batch size one on a single NVIDIA A100 80GB GPU. We discard five warm-up inferences and average over $20$ synchronized runs. We use half-precision inference whenever the released implementation supports it (FP16 autocast for \methodName). ColorfulDiffuse~\cite{Careaga:2024:ColorfulIID} and IntrinsicImageDiffusion~\cite{Kocsis:2024:IntrinsicImageDiffusion} are released in FP32; in our experiments, FP16 did not improve their runtime, so we report their FP32 performance. Since methods predict different numbers of maps, FPS measures the complete output set produced by each method.

\myparagraph{Baseline-specific settings.}
For RGB$\leftrightarrow$X~\cite{Zeng:2024:rgbx}, we use the released checkpoint with DDIM sampling, $50$ denoising steps, one sample per modality, and five separate passes for albedo, irradiance, normals, roughness, and metallicity. IntrinsicImageDiffusion performs inference at its fixed $480{\times}640$ resolution; we use $50$ DDIM steps and average ten independently generated samples, following the paper protocol. DiffusionRenderer~\cite{Liang:2025:DiffusionRenderer} is evaluated in single-image mode with $20$ denoising steps and one pass for each of base color, normals, roughness, metallicity, and depth. MVInverse~\cite{wu:2025:MVInverse} uses a single feed-forward pass; inputs are cropped to patch-compatible dimensions and predictions are resized back to the requested resolution. ColorfulDiffuse is evaluated using the complete stage-4 v2 pipeline with adaptive-resolution processing.

\section{Architecture Details}

We instantiate both teacher and student encoders with DINOv3-Base~\cite{Simeoni:2025:Dinov3} and use the official AdaptFormer design~\cite{Chen:2022:AdaptFormer} in the final six transformer blocks, with bottleneck dimension $384$ and scaling factor $0.1$. The pretrained DINOv3 weights remain frozen throughout training, and only the adapters are optimized. For reconstruction, we follow the official DC-AE decoder architecture~\cite{Chen:2025:DC-AE}, using the DINO feature dimension as the latent channel dimension and an upsampling factor of $16$.

Each modality-specific projection head maps the shared student feature map to the corresponding teacher reconstruction space. We implement each head as a lightweight stack of residual convolutional blocks. Each block consists of a $1{\times}1$ channel projection, a depthwise $3{\times}3$ convolution, GELU activations, and a final $1{\times}1$ projection. The final convolution is zero-initialized, such that each block starts close to the identity mapping and progressively specializes the shared representation for a single intrinsic modality. Each head stacks $K=8$ blocks without normalization layers ($9.51$M parameters per head). We use a separate projection head for albedo, surface normals, irradiance, roughness, and metallicity, while all projected features are decoded by the same DC-AE decoder.

This design keeps the modality-specific transformation lightweight: the backbone provides a common semantic and spatial representation, the projection heads perform feature-space specialization, and the shared decoder maps all modalities back to image space. The full model has $165.45$M parameters: $85.67$M in the frozen DINOv3 backbone, $3.56$M in the adapters, $47.56$M in the five projection heads, and $28.67$M in the decoder.

\begin{table}[t]
\centering
\caption{\textbf{Training configuration.} Both stages use the same six training datasets and keep the DINOv3 backbone frozen.}
\label{tab:training_config}
\footnotesize
\setlength{\tabcolsep}{3pt}
\begin{tabular}{@{}lll@{}}
\toprule
& \textbf{Reconstruction} & \textbf{Decomposition} \\
\midrule
Iterations & 112{,}500 (12 epochs) & 281{,}250 (30 epochs) \\
Global batch & 16 (2 GPUs $\times$ 8) & 16 (8 GPUs $\times$ 2) \\
Input long side & 512 & 512 or 1024, per iteration \\
Optimizer & AdamW, lr $10^{-3}$ & AdamW, lr $10^{-4}$ \\
Schedule & constant, no warm-up & constant, no warm-up \\
Gradient clipping & none & global norm 1.0 \\
Precision & bf16 autocast & bf16 autocast \\
Trainable & adapters, decoder & new adapters, proj.\ heads \\
\bottomrule
\end{tabular}
\end{table}

\section{Training Details}

\Cref{tab:training_config} summarizes the configuration of both stages. AdamW uses $\beta=(0.9, 0.999)$ and no weight decay. An epoch consists of $150{,}000$ samples drawn with replacement: a dataset is chosen uniformly, then a sample is chosen uniformly within it, giving $9{,}375$ iterations per epoch. We use the final checkpoint of each stage, without model selection on test data.

\myparagraph{Reconstruction stage.} Each training sample reconstructs one target chosen at random among its RGB image and its available maps (albedo, normals, irradiance, roughness, metallicity, and depth), using a masked $\ell_1$ loss and LPIPS with unit weights.

\myparagraph{Decomposition stage.} The student adapters are initialized from scratch, while the decoder is copied from the reconstruction stage and kept frozen. For each available map, the feature loss is the channel-mean MSE plus one minus the cosine similarity between the projected student features and the frozen teacher features, computed only on patches with more than $50\%$ valid pixels. The output loss is computed on the decoded map: albedo and irradiance use $\ell_1$, a three-scale gradient loss, and LPIPS in sRGB; normals use the angular error normalized by $\pi$, the gradient loss, and LPIPS; scalar maps (roughness, metallicity, and depth) are obtained by averaging the three decoder output channels before the sigmoid and use $\ell_1$, the gradient loss, and LPIPS. All terms have unit weight, and LPIPS uses AlexNet features. An additional depth head is trained jointly in this stage and discarded afterwards; it is not part of the evaluated model.

\section{Reconstruction Upper Bound}
To separate decoding error from prediction error, we evaluate the reconstruction teacher on ground-truth maps of the test splits used in the main paper: each map is encoded by the teacher and decoded by the shared decoder. \Cref{tab:teacher_upper_bound} compares the result with \methodName{}, which predicts the same features from RGB. The reconstruction space preserves all modalities with high fidelity, so most of the remaining error arises from predicting teacher features from RGB rather than from decoding them. Reconstruction is least accurate for normals on Hypersim ($8.51^\circ$ mean error), suggesting that the $16{\times}$ spatial compression of the decoder also limits fine geometric detail.

\begin{table}[t]
\centering
\caption{\textbf{Reconstruction upper bound.} $X \rightarrow X$ reconstruction of ground-truth maps by the teacher, compared with $\mathrm{RGB} \rightarrow X$ prediction by \methodName{}. Both use the same decoder.}
\label{tab:teacher_upper_bound}
\footnotesize
\setlength{\tabcolsep}{3pt}
\resizebox{\columnwidth}{!}{%
\begin{tabular}{@{}llcccc@{}}
\toprule
& & \multicolumn{2}{c}{\textbf{Hypersim}} & \multicolumn{2}{c}{\textbf{InteriorVerse}} \\
\cmidrule(lr){3-4}\cmidrule(lr){5-6}
\textbf{Map} & \textbf{Metric} & Teacher & Ours & Teacher & Ours \\
\midrule
Albedo & PSNR$\uparrow$ / LPIPS$\downarrow$ & 25.95 / 0.047 & 16.32 / 0.211 & 26.85 / 0.026 & 20.92 / 0.092 \\
Irradiance & PSNR$\uparrow$ / LPIPS$\downarrow$ & 28.32 / 0.044 & 17.86 / 0.231 & -- & -- \\
Normals & Mean$\downarrow$ / Median$\downarrow$ & 8.51 / 4.77 & 16.90 / 9.62 & 6.03 / 2.73 & 8.55 / 2.99 \\
Roughness & PSNR$\uparrow$ / MAE$\downarrow$ & -- & -- & 27.21 / 0.027 & 19.67 / 0.061 \\
Metallicity & PSNR$\uparrow$ / MAE$\downarrow$ & -- & -- & 26.48 / 0.016 & 20.82 / 0.045 \\
RGB & PSNR$\uparrow$ / LPIPS$\downarrow$ & 26.05 / 0.060 & -- & 30.99 / 0.030 & -- \\
\bottomrule
\end{tabular}%
}
\end{table}

\section{Qualitative Results}
\begin{figure*}[t]
    \centering
    \setlength{\parindent}{0pt}
    \setlength{\parskip}{0pt}

    \intrinsiclabelrow\par\vspace{1pt}
    \includegraphics[width=\textwidth]{images/room_intrinsic_out.jpg}\par\vspace{0.1mm}
    \includegraphics[width=\textwidth]{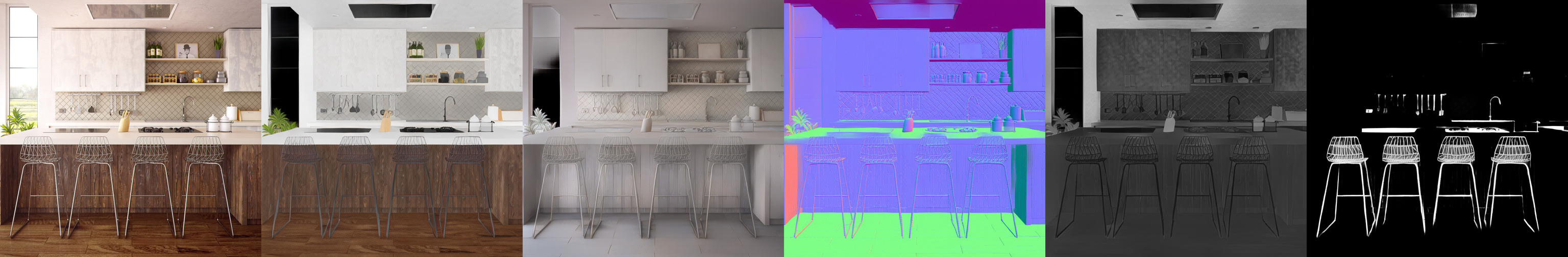}\par\vspace{0.1mm}
    \includegraphics[width=\textwidth]{images/paris_church_intrinsic_out.jpg}\par\vspace{0.1mm}
    \includegraphics[width=\textwidth]{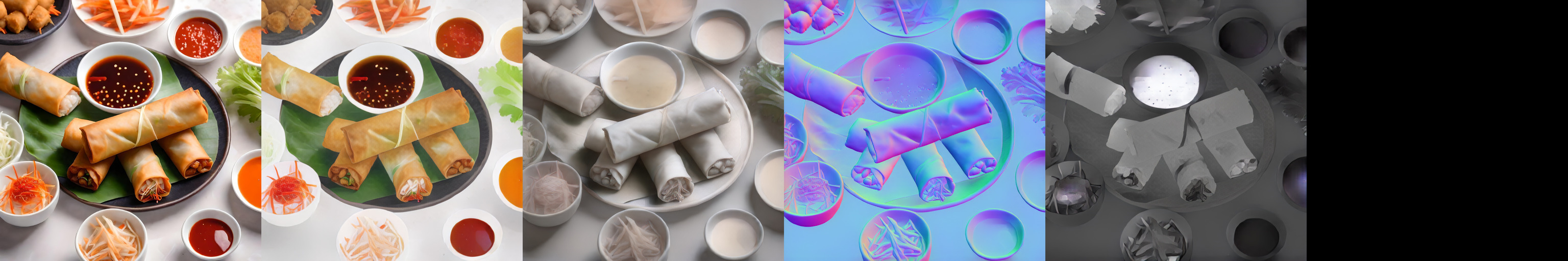}\par\vspace{0.1mm}
    \includegraphics[width=\textwidth]{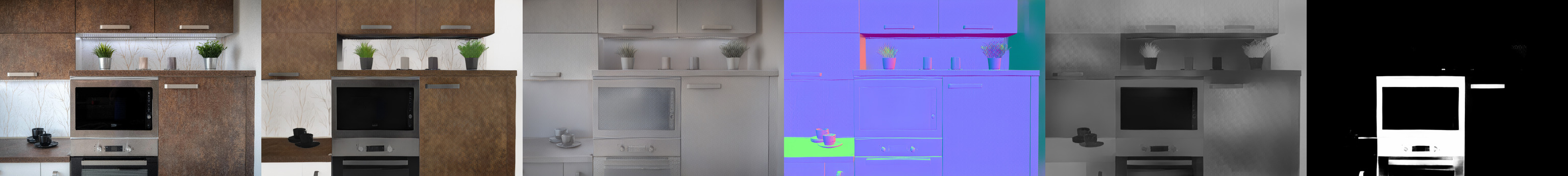}\par\vspace{0.1mm}
    \includegraphics[width=\textwidth]{images/street_intrinsic_out.jpg}\par\vspace{0.1mm}
    \includegraphics[width=\textwidth]{images/table_ice_cream_intrinsic_out.jpg}\par\vspace{0.1mm}
    \includegraphics[width=\textwidth]{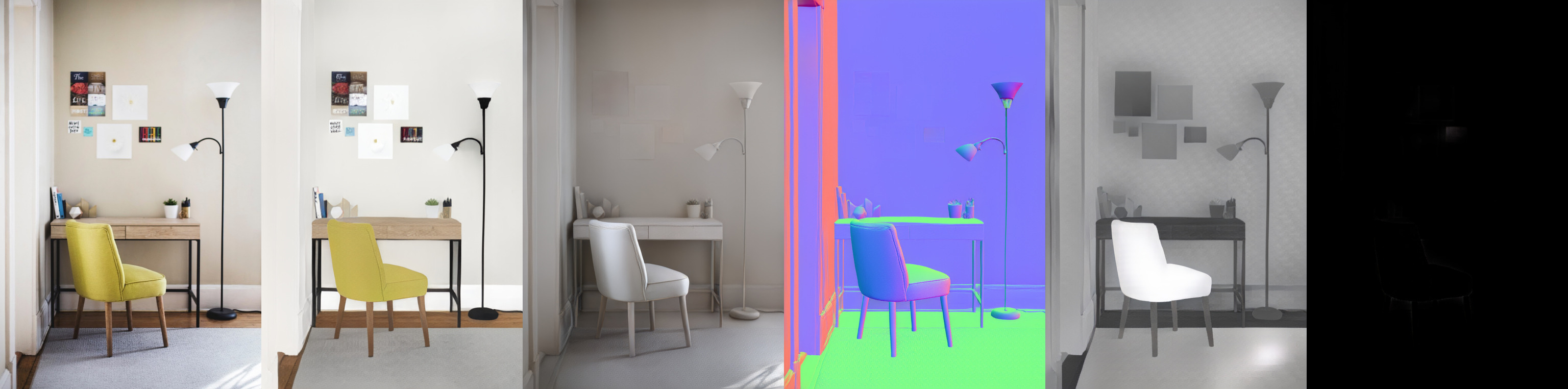}\par\vspace{0.1mm}

    \caption{\textbf{Additional qualitative results.} Intrinsic maps predicted by \methodName{} for additional input images.}
    \label{fig:intrinsic_results}
\end{figure*}

\Cref{fig:intrinsic_results} presents qualitative predictions across diverse scenes. Despite using a single shared model to estimate all intrinsic properties jointly, rather than separate networks specialized for individual outputs, the predictions are coherent across appearance, illumination, geometry, and materials. In particular, albedo, irradiance, and surface normals remain visually consistent across a broad range of objects and scene layouts. Roughness and metallicity are the most challenging modalities, as they are often only weakly constrained by appearance and can be confused with color, shading, or specular highlights. In the penultimate example, the cherry stem is incorrectly predicted as metallic. In one of the kitchen scenes, several cabinets have visually similar handles, but only one is correctly identified as metallic while the others are missed. Overall, the qualitative results show that a single compact feed-forward model can recover a broad set of intrinsic properties jointly, while maintaining strong spatial detail and coherent predictions across modalities.

\end{document}